%% file: main.tex
\documentclass{article}

\usepackage{arxiv}
\usepackage[utf8]{inputenc}
\usepackage[T1]{fontenc}
\usepackage{microtype}
\usepackage{graphicx}
\usepackage{booktabs}
\usepackage{array}
\usepackage{tabularx}
\usepackage{amsmath}
\usepackage{amssymb}
\usepackage{xcolor}
\usepackage{placeins}
\usepackage{float}
\usepackage{hyperref}

\hypersetup{
  colorlinks=true,
  linkcolor=blue,
  citecolor=blue,
  urlcolor=blue,
  pdftitle={Does Super-Resolution Preserve Defect Evidence? A Low-False-Call Benchmark for Semiconductor Inspection},
  pdfauthor={Shaoliang Yang and Jun Wang},
  pdfsubject={Task-aware evaluation of super-resolution for semiconductor defect inspection},
  pdfkeywords={semiconductor inspection, super-resolution, defect detection, low false-call rate, operating-point transfer, calibration, benchmark}
}

\title{Does Super-Resolution Preserve Defect Evidence?\\
A Low-False-Call Benchmark for Semiconductor Inspection}
\author{%
Shaoliang Yang\\
Santa Clara University\\
Santa Clara, CA 95053, USA\\
\texttt{syang11@scu.edu}
\And
Jun Wang\thanks{Corresponding author: \texttt{jwang22@scu.edu}}\\
Santa Clara University\\
Santa Clara, CA 95053, USA\\
\texttt{jwang22@scu.edu}
}
\date{}
\renewcommand{\headeright}{arXiv Preprint}
\renewcommand{\undertitle}{arXiv Preprint}
\renewcommand{\shorttitle}{Defect-Evidence Preservation Under Super-Resolution}

\newcommand{\fpr}{\mathrm{FPR}}
\newcommand{\nhr}{\mathrm{NHR}}

\newcommand{\tabref}[1]{\hyperref[#1]{Table~\ref*{#1}}}
\newcommand{\figref}[1]{\hyperref[#1]{Fig.~\ref*{#1}}}
\newcommand{\secref}[1]{\hyperref[#1]{Section~\ref*{#1}}}
\newcommand{\secrefs}[2]{\hyperref[#1]{Sections~\ref*{#1}--\ref*{#2}}}
\newcommand{\appref}[1]{\hyperref[#1]{Appendix~\ref*{#1}}}

\newcounter{algorithmctr}
\newenvironment{algobox}[1]{%
  \refstepcounter{algorithmctr}%
  \par\medskip%
  \noindent\begin{minipage}{\linewidth}%
  \noindent\hrule height 0.8pt\relax%
  \vspace{0.18em}%
  \noindent\textbf{Algorithm \thealgorithmctr\quad #1}\par%
  \vspace{0.12em}%
  \noindent\hrule height 0.4pt\relax%
}{%
  \vspace{0.18em}%
  \noindent\hrule height 0.4pt\relax%
  \end{minipage}%
  \par\medskip%
}

\begin{document}
\maketitle

\begin{abstract}
Super-resolution can make inspection images appear sharper without preserving the evidence needed to detect a defect. We study this failure mode with a benchmark that separates reconstruction from detection and evaluates both at a predeclared low false-positive rate. Ten end-to-end repetitions combine independently generated line/space and contact-hole images with model training, calibration, clean controls, weak defects, and a held-out defect morphology. Every reconstruction is scored by the same local residual detector, while direct and jointly trained detectors form a separate comparison track. Reconstruction fidelity and inspection utility diverge: the two learned reconstruction models attain the highest structural similarity yet detect fewer defect pixels than bicubic interpolation in every paired repetition. A direct DeepLabV3 detector reaches $0.1984\pm0.0385$ recall at $0.000174\pm0.000084$ false-positive rate and satisfies the held-out feasibility criterion in all ten repetitions. An illustrative joint model, DPU-WaferSR, passes independent clean calibration but exceeds the held-out limit in all ten repetitions, demonstrating that calibration success does not guarantee transfer. Weak-defect recall remains near zero for every feasible method. Applying the unchanged policies to 4,591 public Carinthia-S masks further reveals large method-dependent shifts on real SEM texture. These results support a simple conclusion: super-resolution for inspection should be judged by preserved task evidence and operating-point transfer, not reconstruction quality alone.
\end{abstract}

\input{sections/01_introduction}
\input{sections/02_related_work}
\input{sections/03_problem_benchmark}
\input{sections/04_methods_protocol}

\input{sections/05_results}

\input{sections/06_discussion}

\input{sections/07_appendices}

\bibliographystyle{unsrt}
\begingroup
\small
\bibliography{references}
\endgroup

\end{document}

%% file: sections/01_introduction.tex
\section{Introduction}
\label{sec:introduction}

Semiconductor inspection images are acquired to support a decision: identify a suspect location, route it for review, and decide whether the observed structure is consistent with the process. Super-resolution (SR) is attractive in this setting because it promises a higher-resolution view without changing the acquisition system. The usual evidence for improvement---peak signal-to-noise ratio (PSNR), structural similarity (SSIM), or visual sharpness---does not answer the operational question. A reconstruction can reproduce the dominant periodic pattern while suppressing a narrow bridge, filling a small gap, inventing an edge on a clean line, or moving a boundary enough to change a thresholded decision. These errors can be inconspicuous in an average reconstruction loss because the defect occupies only a small fraction of the image.

The distinction between image quality and scientific evidence is well recognized in microscopy reconstruction. Learned restoration can recover useful signal from low-SNR observations, but a plausible-looking image is not automatically a faithful measurement \cite{belthangady2019microscopy,weigert2018care,qiao2021deepsem}. The distinction is especially consequential in wafer inspection. Reviews of SEM inspection emphasize defect discovery, localization, and review burden rather than visual preference alone \cite{sem_review2023}. Recent wafer-specific SR studies span confocal microscopy, scanning acoustic microscopy, optical nanoscale inspection, and TEM reconstruction \cite{sun2025confocal_sr,wilhelmer2025ai_enhancement,srfabnet2025,kim2026wafer_tem_sr}. They demonstrate the promise of faster or task-aware acquisition, while also showing why reconstructed images require domain-specific validation. Modern SR architectures have advanced from residual convolutional networks to channel attention, blind-degradation models, transformers, and simplified gated restoration blocks \cite{lim2017edsr,zhang2018rcan,wang2021realesrgan,liang2021swinir,chen2023hat,chen2022nafnet}. Their standard restoration metrics, however, do not establish whether a downstream inspection decision remains reliable at a low false-call burden.

The operating constraint changes the interpretation of a recall gain. In a sparse-defect problem, a method can increase recall simply by activating more often. If this behavior increases clean-region calls beyond the review budget, the nominal gain is not an inspection improvement at the intended operating point. Conversely, a model can appear safe by remaining nearly silent while missing the weak evidence that motivates higher-resolution imaging. Precision--recall analysis makes the rare-positive tradeoff explicit \cite{davis2006pr}, but a curve alone does not test whether a threshold selected before evaluation transfers to held-out clean structure. Inspection-oriented evaluation therefore requires both a declared false-positive target and a strict separation among the data used for model fitting, probability calibration, operating-point selection, and final testing.

SR adds a second causal difficulty. If each reconstruction is scored with a different detector or detector hyperparameter, changes in reconstruction and changes in scoring are confounded. Track~A therefore fixes the scoring function before comparing image transformations. Every image-producing method is evaluated with the same detector, the same local-residual smoothing parameter, the same calibration targets, and the same metric implementation. Bicubic interpolation is the single reference used to place the low-resolution observation on the high-resolution grid. Under this design, a paired change in detector response can be attributed to the transformed image rather than to method-specific scoring.

Direct mask predictors answer a different question and are therefore kept in a separate track. A direct detector can exploit evidence in the low-resolution input without producing an image whose reconstruction fidelity is meaningful. Its performance demonstrates how much task evidence is available, but it cannot be interpreted as an SR gain. Likewise, a joint reconstruction/detection model can test whether clean calibration transfers, yet its learned defect head prevents a causal comparison with image-only transformations. Separating these tracks turns one broad model leaderboard into two interpretable experiments: whether an image transformation preserves evidence for a fixed detector, and whether a task-trained predictor can recover evidence that reconstruction-only training leaves unused.

\begin{figure}[!htbp]
  \centering
  \includegraphics[width=\linewidth]{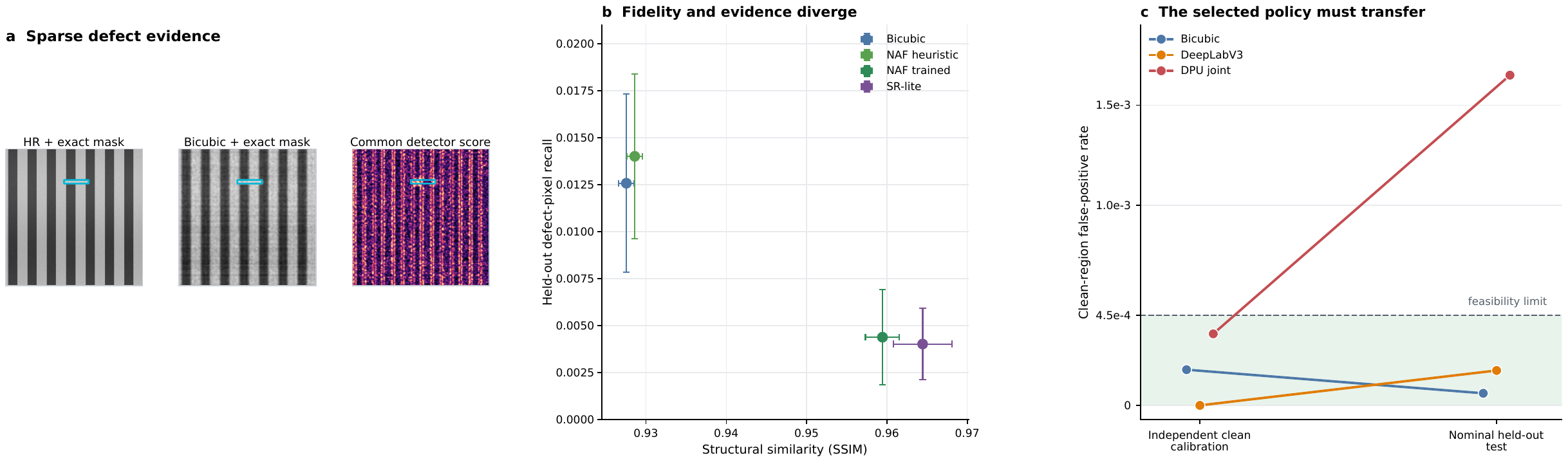}
  \caption{Why inspection-oriented SR requires task evidence and operating-point transfer. (a) A representative bridge occupies a small region within a periodic line/space image; the cyan contour marks the exact defect mask, and the common detector acts on a local residual rather than visual sharpness. (b) Ten-run means show that the learned reconstruction models can obtain higher SSIM and lower held-out defect recall than bicubic interpolation; bars show one standard deviation. (c) The same preselected policy can behave differently on independent clean calibration and nominal held-out images. The green band lies below the predeclared $4.5\times10^{-4}$ feasibility limit. Complete method and seed-level results are reported in \secref{sec:results}.}
  \label{fig:longform_overview}
\end{figure}

\figref{fig:longform_overview} previews the resulting argument. The exact mask occupies a small region inside a dominant periodic pattern. In the reconstruction track, higher SSIM does not imply higher defect recall. In the operating-policy experiment, a policy that satisfies independent clean calibration can still exceed the same false-positive limit on nominal held-out images. The complete figures later in the paper show all methods and all ten repeated policies; the opening figure is intended to make the two scientific failure modes visible before the protocol is developed.

We introduce WaferInspectSR-Bench to support this controlled test. The synthetic layer provides paired $128\times128$ low-resolution and $256\times256$ high-resolution images, exact defect masks, defect-edge masks, clean-region masks, clean-only controls, attenuated weak defects, and a residue morphology withheld from nominal training. The generator covers line/space and contact-hole layouts with bridge, gap, missing-pattern, particle, and scratch defects. Its degradations include blur, defocus, twofold downsampling, shot/Poisson/Gaussian noise, scan-line structure, contrast drift, and aliasing. These factors form a reproducible imaging stress model; they are not presented as a fitted model of a specific proprietary scanner.

The external layer uses all 4,591 public Carinthia-S image/mask pairs \cite{carinthias2025}. Each external image is converted to a pseudo-low-resolution observation, and the synthetic policies are applied without retraining, probability refitting, threshold selection, or prior-weight adjustment. Because Carinthia-S does not provide paired optical acquisitions at two resolutions, the external experiment is not a reconstruction benchmark or a fab-deployment validation. It is a deliberately difficult transfer test: a policy that appears quiet on controlled synthetic images should not be assumed to remain quiet on real SEM texture.

The empirical study uses ten predeclared end-to-end repetitions. Each repetition regenerates the synthetic images and degradations, trains the neural participants, fits a probability temperature on validation images, selects among candidate thresholds using independent clean images, and evaluates all held-out partitions once. The repeated unit is the complete seed-level pipeline rather than an individual pixel. This choice avoids treating millions of spatially correlated pixels as independent observations and permits paired method comparisons on the same generated realization, following the paired-comparison logic used in statistical machine-learning evaluation \cite{demsar2006statistical}.

The paper makes four contributions:

\begin{enumerate}
  \item \textbf{A controlled reconstruction experiment.} Interpolation, classical restoration, a deterministic NAF-style enhancement, and two trained reconstruction networks are evaluated through one identical local-residual detector with smoothing parameter $\sigma=2.0$. Only the transformed image changes.
  \item \textbf{A predeclared operating-point transfer procedure.} Probability temperature and candidate thresholds are fitted on validation images. The most permissive candidate satisfying the false-positive limit is selected on independent clean calibration images. Nominal, weak, clean, morphology-shift, and external images are then evaluated once without post-test adjustment.
  \item \textbf{A separated task-trained comparison.} A direct DeepLabV3 detector establishes the task-evidence regime of the low-resolution observation, while a compact joint reconstruction/detection model illustrates how a selected clean-calibration policy can fail to transfer. These models are not conflated with reconstruction effects.
  \item \textbf{Complete task-conditioned reporting.} The paper reports pixel, component, boundary, reconstruction, clean-only, weak-defect, morphology-shift, and external metrics; seed-level paired differences; and exact transformation and training definitions needed to interpret the numerical results.
\end{enumerate}

Three findings organize the story. First, SR-lite and the trained NAF-style reconstruction attain the highest SSIM but lower defect recall than bicubic interpolation in every paired repetition. Second, DeepLabV3 recovers substantially more nominal defect evidence at a feasible false-positive rate, whereas the illustrative joint model satisfies clean calibration and fails the held-out limit in all ten repetitions. Third, weak defects remain nearly undetected for every feasible method, and real SEM texture causes large, method-dependent changes in false-call behavior. The result is not a claim that SR has no value. It is evidence that reconstruction quality alone does not establish inspection value, and that an operating threshold is a policy whose transfer must be measured.

The remainder develops the argument from prior work and problem definition through protocol, controlled results, interpretation, and limitations. \secref{sec:related} positions the study against restoration, scientific imaging, anomaly localization, task-aware wafer SR, and calibration work. \secref{sec:problem} defines evidence preservation and feasibility. \secref{sec:benchmark} describes the generator, masks, degradation, and external layer. \secrefs{sec:methods}{sec:protocol} specify the transformations, direct and joint models, common detector, selection algorithm, metrics, and repeated-unit analysis. \secref{sec:results} reports the controlled results. \secref{sec:discussion} interprets the findings, and \secref{sec:limitations} states the scope boundaries. The appendices provide model profiles, training details, seed-level summaries, count checks, and metric definitions.

%% file: sections/02_related_work.tex
\section{Related Work and Study Positioning}
\label{sec:related}

The closest literature spans several communities that use different definitions of success. Image SR emphasizes reconstruction or perceptual quality; scientific imaging adds measurement validity; industrial anomaly detection emphasizes fixed masks and localization; semiconductor inspection adds low false-call and review-burden constraints; and calibration research asks whether a selected decision rule remains reliable. The present study lies at their intersection. Its contribution is an evaluation design rather than a new SR architecture.

\subsection{Reconstruction-centered super-resolution}

Single-image SR has a long reconstruction-centered lineage. SRCNN introduced an early convolutional mapping from low-resolution to high-resolution image patches \cite{dong2014srcnn}. VDSR deepened the residual-learning formulation \cite{kim2016vdsr}, while LapSRN used a coarse-to-fine pyramid to improve efficiency and reconstruction quality \cite{lai2017lapsrn}. EDSR removed unnecessary normalization from residual blocks and established a strong PSNR-oriented baseline \cite{lim2017edsr}. RDN and RCAN added dense feature reuse and channel attention, respectively, to recover high-frequency content more effectively \cite{zhang2018rdn,zhang2018rcan}. These models are important reference families because they show how strongly a network can optimize average reconstruction error. That same strength creates the inspection question: if periodic background pixels dominate the loss, does the network retain the sparse pixels that determine a defect call?

Transformer and simplified restoration designs have broadened the architecture space. IPT uses large-scale pretraining for restoration \cite{chen2021ipt}; SwinIR adapts shifted-window transformers to denoising, deblocking, and SR \cite{liang2021swinir}; HAT combines channel attention and self-attention for high-capacity SR \cite{chen2023hat}; and NAFNet shows that a simplified gated block can be competitive without conventional nonlinear activations \cite{chen2022nafnet}. These advances motivate the NAF-style participants used here, but architecture sophistication is not the independent variable in our primary experiment. The reconstruction track holds the detector fixed and asks whether the transformed image preserves more measurable evidence.

RealSR is a particularly relevant benchmark precedent because it shows that bicubic synthesis alone is insufficient for real-camera SR and makes cross-device transfer part of evaluation \cite{cai2019realsr}. Blind and real-world methods such as BSRGAN and Real-ESRGAN similarly broaden the degradation family to improve robustness on natural images \cite{zhang2021bsrgan,wang2021realesrgan}. Our work follows the principle that acquisition assumptions must be explicit and transfer must be measured. It differs in the target variable: low-false-call defect evidence, rather than perceptual natural-image quality, is the primary outcome.

\subsection{Perceptual restoration and hallucination risk}

GAN-based methods made the conflict between distortion and perceptual realism visible. SRGAN and ESRGAN can produce sharper and more plausible textures than distortion-oriented models \cite{ledig2017srgan,wang2018esrgan}. Diffusion-based SR extends this generative view \cite{saharia2023sr3}. Such outputs can be appropriate for visualization, but plausibility is not equivalent to provenance. A texture synthesized because it is statistically likely can resemble a physical defect; a smooth reconstruction can remove an unlikely but real structure. In a thresholded inspection system, both events become operational errors.

This concern does not imply that generative SR is intrinsically unsuitable for scientific images. It implies that the output must be validated against the task and that the original observation should remain available. Inspection evidence is not fully characterized by pixel agreement, yet a task score alone can also be gamed by broad activation. The combination of a common detector, explicit clean controls, exact masks, and a predeclared operating point is intended to expose both forms of failure.

\subsection{Scientific microscopy and semiconductor imaging}

Scientific-image restoration demonstrates that learned models can recover useful information under constrained acquisition. CARE denoises and restores fluorescence microscopy from low-SNR observations \cite{weigert2018care}. Reviews of deep learning for microscopy SR emphasize improvements in acquisition speed and resolution while also stressing the need for domain validation \cite{belthangady2019microscopy}. Learning-based optical-microscopy reconstruction and sparse SEM acquisition studies similarly seek to reduce acquisition burden or improve apparent detail \cite{wang2019deepsem,trampert2019sparse_sem,qiao2021deepsem}. These studies motivate restoration as an instrument-level tool, but most do not impose a rare-event false-positive constraint with clean-only controls.

Semiconductor inspection makes that constraint central. SEM inspection reviews describe heterogeneous defect morphologies, limited labels, structured background, and throughput pressure \cite{sem_review2023}. Public descriptions of inspection and review systems distinguish high-throughput discovery from higher-resolution review and disposition \cite{kla_defect_review,kla_bbp_39xx,kla_3900_press2016}. A false call consumes review capacity; a missed weak defect can hide process drift. The benchmark therefore measures both defect recall and the clean-region burden at the same selected policy. It does not model proprietary scanner optics or claim production throughput.

Wafer-specific SR now spans confocal microscopy for faster optical acquisition, scanning acoustic microscopy coupled to segmentation and detection, optical SR-guided defect detection, and TEM reconstruction \cite{sun2025confocal_sr,wilhelmer2025ai_enhancement,srfabnet2025,kim2026wafer_tem_sr}. The wafer-TEM study is especially close: it evaluates reconstructed inputs while keeping a pretrained SAM evaluator fixed, and it notes that artifacts can generate false positives or obscure true structures \cite{kim2026wafer_tem_sr}. We therefore do not claim novelty for wafer SR, downstream evaluation, or fixed-model evaluation by itself. The contribution here is the inspection-specific control: image transformations are scored under a preselected low-FPR policy, independent clean selection is separated from no-retune held-out evaluation, and clean, weak-defect, morphology-shift, and external transfer tests remain outside policy selection.

\subsection{Defect detection and anomaly localization}

Dense prediction provides an alternative to reconstruction. FCN and U-Net established practical pixelwise segmentation architectures \cite{long2015fcn,ronneberger2015unet}; DeepLabV3 uses atrous convolution and multiscale context to improve semantic localization \cite{chen2017deeplabv3}. These models are relevant because a direct detector can optimize the sparse mask rather than an average image loss. In our study, DeepLabV3 serves as an evidence-availability comparator: it asks whether the degraded observation contains task signal that a reconstruction-only objective fails to preserve.

Industrial anomaly benchmarks contribute a complementary protocol tradition. MVTec AD and MVTec AD~2 provide public images, anomaly masks, and standardized evaluation for industrial localization \cite{bergmann2019mvtec,bergmann2026mvtecad2}. PaDiM models the distribution of normal features at each spatial location and detects deviations without defect examples \cite{defard2021padim}. These benchmarks begin after the image has been acquired. WaferInspectSR-Bench inserts an explicit reconstruction intervention before localization, so the effect of that intervention can be studied.

Carinthia and Carinthia-S provide public SEM images and masks that make external stress testing possible \cite{carinthia2024,carinthias2025}. They do not supply paired high- and low-resolution optical acquisitions, and their class distribution is highly imbalanced. We therefore use Carinthia-S only at inference, report both image-macro and class-macro recall, and avoid interpreting the external result as an SR reconstruction score.

\subsection{Task-aware and SR-guided defect models}

Task-aware SR replaces or augments reconstruction loss with a downstream objective. Task-driven detection and SR4IR establish this general lineage for object detection, segmentation, and classification \cite{haris2021task_driven_sr,kim2024sr4ir}. GeoSR-Bench extends the same principle to benchmark design and shows that fidelity rankings need not predict downstream-task rankings in remote sensing \cite{li2026geosr_bench}. We therefore do not claim novelty for task-driven SR or for the general observation that fidelity and task utility can diverge.

SR-FABNet is the closest joint-model precedent in optical wafer inspection: it couples an SR branch to a Fourier-attention defect detector \cite{srfabnet2025}. Its focused architecture story motivates joint modeling, whereas our contribution is a benchmark-level causal control. An image-only SR method and a direct detector cannot be placed in one undifferentiated ranking. The common-detector reconstruction track measures the intervention of interest; direct and joint models occupy a second track that reveals the attainable task regime and calibration behavior. The specific added question is whether one fixed detector and a preselected low-FPR policy preserve their behavior after reconstruction and distribution transfer.

A definitive task-aware architecture comparison would hold backbone, parameter count, training data, and optimization budget fixed while changing only the task supervision. The present study does not claim to complete that comparison. It establishes why it is needed: the reconstruction-only models improve SSIM while losing paired defect recall, and the direct detector shows that more nominal task evidence is available. This result turns a future architecture ablation into a motivated experiment rather than an arbitrary extension.

\subsection{Calibration, selection, and repeated comparisons}

Temperature scaling can improve the calibration of neural probabilities with a single fitted parameter \cite{guo2017calibration}. Expected calibration error and related measures summarize the mismatch between confidence and empirical frequency \cite{naeini2015calibration}. Calibration, however, does not guarantee control after a shift in image content. A threshold that appears safe on defect-free calibration images may activate more often on nominal images containing boundaries, or on external SEM texture. The policy itself must therefore be tested after selection.

Low-FPR evaluation is sensitive to data leakage because a threshold can be made to look feasible by inspecting the test false-positive rate and choosing a stricter candidate. We use three stages instead: validation images fit temperature and candidate thresholds; independent clean calibration images select the most permissive feasible candidate; held-out images are evaluated once. A policy that fails after this sequence is not retuned. Its failure is evidence about constraint transfer.

The unit of analysis also matters. Pixelwise confidence intervals can become implausibly narrow when spatially correlated pixels are treated as independent. We first aggregate metrics within each complete seed-level pipeline and then summarize the ten repetitions. Paired differences compare methods on the same generated data, consistent with paired algorithm-comparison procedures in statistical machine-learning evaluation \cite{demsar2006statistical}. Pooled-pixel counts remain useful descriptive checks, but they do not replace seed-level dispersion. \tabref{tab:peer_positioning} summarizes how this design differs from the closest restoration, anomaly-localization, and wafer-inspection precedents.

\begin{table}[!htbp]
\centering
\caption{Relationship to close peer families. The last column states the evaluation question added here rather than implying that one benchmark subsumes another.}
\label{tab:peer_positioning}
\small
\begin{tabularx}{\linewidth}{>{\raggedright\arraybackslash}p{0.18\linewidth}>{\raggedright\arraybackslash}p{0.23\linewidth}>{\raggedright\arraybackslash}p{0.27\linewidth}X}
\toprule
Study family & Primary object & Principal strength & Question added here \\
\midrule
RealSR \cite{cai2019realsr} & real-image SR & paired observations and cross-camera transfer & task evidence under a low false-call limit \\
MVTec AD / AD~2 \cite{bergmann2019mvtec,bergmann2026mvtecad2} & industrial anomaly localization & fixed public data and mask evaluation & the effect of a preceding SR transformation \\
Task-driven SR / SR4IR \cite{haris2021task_driven_sr,kim2024sr4ir} & downstream recognition & task-supervised reconstruction & independent low-FPR policy selection and transfer \\
GeoSR-Bench \cite{li2026geosr_bench} & downstream SR benchmarking & multi-model, multi-task fidelity--utility analysis & rare-event inspection controls and frozen-policy transfer \\
Wafer confocal / acoustic SR \cite{sun2025confocal_sr,wilhelmer2025ai_enhancement} & inspection imaging and throughput & real acquisition and downstream analysis workflows & causal detector control under a shared operating policy \\
SR-FABNet \cite{srfabnet2025} & SR-guided wafer detection & focused task-aware joint model & causal separation of reconstruction and detector effects \\
Wafer-TEM SR \cite{kim2026wafer_tem_sr} & microscopy-specific SR & domain reconstruction and downstream analysis & independent clean selection and held-out false-call transfer \\
WaferInspectSR-Bench & inspection evidence preservation & common-detector reconstruction track, direct-model track, weak and clean controls & whether a sharper reconstruction improves a feasible inspection decision \\
\bottomrule
\end{tabularx}
\end{table}

The resulting scope is deliberately bounded. The benchmark does not propose a production scanner model, a universally best detector, or a perceptual-quality replacement. It supplies an exact-mask environment in which the relationship among reconstruction fidelity, preserved task evidence, clean-region burden, and policy transfer can be tested without changing more than one causal element at a time.

%% file: sections/03_problem_benchmark.tex
\section{Problem Formulation}
\label{sec:problem}

Let $x\in[0,1]^{H\times W}$ be a high-resolution inspection image, $y=\mathcal{D}(x)$ its low-resolution observation, $M\in\{0,1\}^{H\times W}$ the defect mask, $C\in\{0,1\}^{H\times W}$ the clean-region mask, and $E\in\{0,1\}^{H\times W}$ the defect-edge mask. An image transformation $g$ produces $z=g(y)$, while a direct or joint model may produce a defect score $p$ and, optionally, a reconstructed image $\widehat{x}$ and risk score $r$. A threshold $\tau$ yields the binary prediction
\begin{equation}
\widehat{M}_{\tau}(u)=\mathbb{1}[p(u)\geq\tau],
\end{equation}
where $u$ indexes output pixels.

The primary objective is intentionally narrow:
\begin{equation}
\max \; \mathrm{Recall}(\widehat{M}_{\tau},M)
\quad\text{subject to}\quad
\fpr(\widehat{M}_{\tau},C)\leq\alpha,
\label{eq:primary_objective}
\end{equation}
with target $\alpha=3\times10^{-4}$. Finite calibration samples make exact equality to a rare-event target brittle, so a predeclared tolerance defines held-out feasibility as $\fpr\leq4.5\times10^{-4}$. This tolerance is a research-analysis rule, not a fab acceptance limit. It is held constant for all methods.

\begin{figure}[!htbp]
  \centering
  \includegraphics[width=\linewidth]{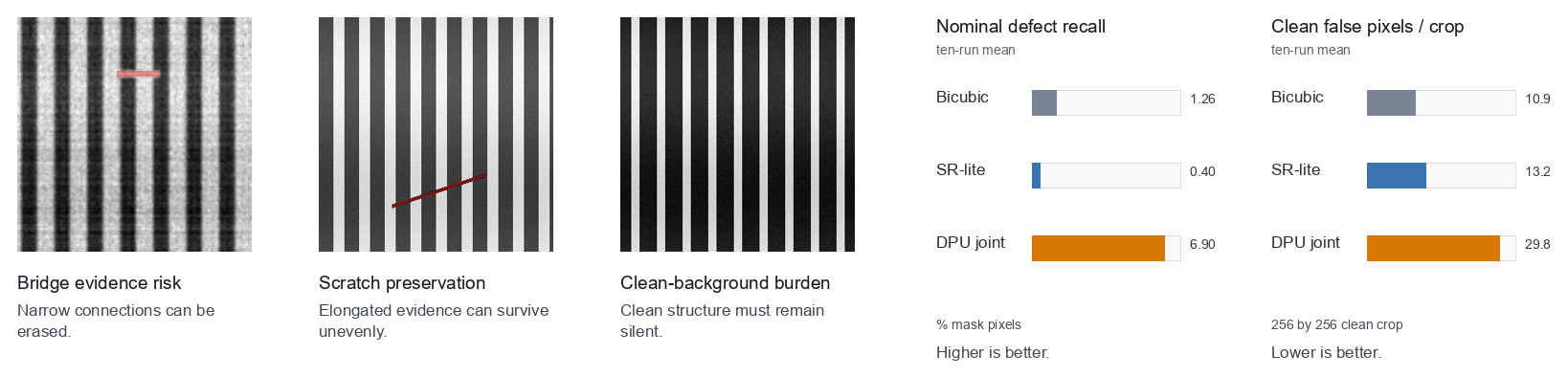}
  \caption{Failure modes and representative current outcomes for inspection-oriented SR evaluation. The image panels show how a reconstruction can erase a narrow bridge, preserve an elongated defect unevenly, or activate a clean region. The bars report ten-run means at the frozen policies for nominal defect-pixel recall and false pixels per clean $256\times256$ crop. DPU-WaferSR is the separate joint-model stress participant, not a Track~A reconstruction method.}
  \label{fig:motivation}
\end{figure}

Equation~\eqref{eq:primary_objective} distinguishes evidence preservation from broad activation. A method improves the primary outcome only when additional recovered defect pixels remain compatible with the same clean-region burden. Secondary outcomes reported in the main comparisons include SSIM for image-producing methods, precision, edge F1, component recall, component IoU, and clean-only hallucination; they explain why two policies with similar primary metrics may behave differently, but they are not undeclared acceptance constraints. \figref{fig:motivation} summarizes the corresponding failure modes and reports the current ten-run recall and clean false-call burden for three representative frozen policies.

\subsection{Image evidence, detector evidence, and task evidence}

Three objects must be kept separate. \emph{Image evidence} is the structure present in $z$ after interpolation or reconstruction. \emph{Detector evidence} is the score produced by a fixed rule applied to that image. \emph{Task evidence} is the score learned directly from the low-resolution input. The reconstruction experiment changes $g$ while holding the downstream scoring rule constant. The direct-model experiment changes the scoring model and therefore answers a different question.

For Track~A, evidence preservation is a property of the composite $h\circ g$, where $h$ is one common local-residual detector. Because $h$ is identical for every transformation, a paired change in recall can be attributed to how $g$ modifies the signal used by $h$. The study does not claim that $h$ is the optimal detector for every reconstruction; method-specific detector tuning would answer a system-design question rather than isolate the reconstruction.

For Track~B, a direct model $d_{\phi}(y)$ learns the defect score without reconstructing the image. Its performance establishes whether usable mask information exists in $y$ under task supervision. The joint model $f_{\theta}(y)=(\widehat{x},p,r)$ provides another system-level participant. These models can be compared with one another at the same operating-policy procedure, but their result cannot be interpreted as the causal effect of SR.

\subsection{Operating policy and transfer}

The evaluated policy contains the fitted model weights, probability temperature, threshold, and any fixed postprocessing or prior-fusion setting. Training images update weights. Validation calibration images fit the temperature and the set of candidate thresholds. Independent clean calibration images select among those candidates. The selected policy is then applied unchanged to nominal, clean, weak-defect, morphology-shift, and external images.

This ordering produces two distinct feasibility statements. \emph{Selection feasibility} means the independent clean calibration images satisfy the limit. \emph{Held-out feasibility} means the nominal test images satisfy the same limit after selection. A selection-feasible policy may fail on held-out data because the clean calibration set and the nominal background do not have identical structure. The gap is a measured transfer outcome rather than a reason to choose a new threshold. \tabref{tab:notation} collects the image, mask, score, and policy symbols used throughout the comparison.

\begin{table}[!htbp]
\centering
\caption{Core notation and its role in the controlled comparison.}
\label{tab:notation}
\small
\begin{tabular}{lll}
\toprule
Symbol & Object & Role \\
\midrule
$x,y,z$ & HR target, LR observation, transformed image & reconstruction path \\
$M,C,E$ & defect, clean-region, and edge masks & task labels and clean accounting \\
$p,r$ & defect score and optional risk score & thresholding and selective analysis \\
$g,h,d_{\phi}$ & image transformation, common detector, direct detector & separated causal roles \\
$\tau,T$ & threshold and scalar temperature & selected operating policy \\
$\alpha$ & target clean-region FPR & primary operating constraint \\
$\widehat{M}_{\tau}$ & thresholded mask & pixel and component evaluation \\
\bottomrule
\end{tabular}
\end{table}

\section{WaferInspectSR-Bench}
\label{sec:benchmark}

\begin{figure}[!htbp]
  \centering
  \includegraphics[width=\linewidth]{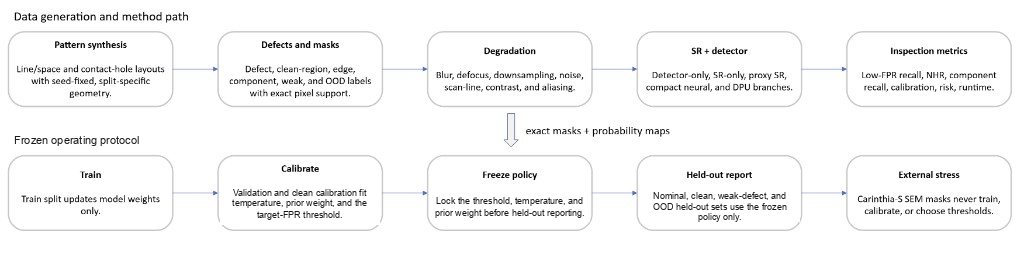}
  \caption{End-to-end data and protocol path. Exact masks make low-FPR, clean-region, edge, component, and calibration metrics traceable. Thresholds, temperature, and prior weight are frozen before held-out and external reporting; Carinthia-S masks never enter training or operating-point selection.}
  \label{fig:pipeline}
\end{figure}

The benchmark has a controlled synthetic layer and an external public SEM layer. The controlled layer supplies paired images and exact masks for causal evaluation. The external layer tests whether unchanged synthetic policies remain sensible on real texture. The two layers are reported separately because they support different inferences. \figref{fig:pipeline} shows the end-to-end data and protocol path: exact masks support the low-FPR and clean-region metrics, operating-policy parameters are frozen before held-out reporting, and Carinthia-S masks never enter training or threshold selection.

\subsection{Synthetic layouts, defects, and split design}

The generator produces line/space and contact-hole layouts at $256\times256$ resolution. Nominal defect families are bridges, gaps, missing-pattern regions, particles, and scratches. Weak cases attenuate those nominal defects. Residue is an irregular morphology withheld from nominal fitting and selection. Clean-only images contain no defect and exist specifically to measure hallucination without a nearby true boundary.

Each of ten predeclared seeds regenerates the same split schema with different layout, defect, and degradation draws. The seeds are 7, 11, 13, 17, 19, 23, 29, 31, 37, and 41. Each pipeline contains 304 images, giving 3,040 synthetic samples across the study. The training partition contains 128 nominal defect images. Thirty-two validation images fit temperature and candidate thresholds. Sixteen independent no-defect images select the operating point. The held-out partitions contain 48 nominal defects, 24 clean controls, 24 weak defects, and 24 residue defects. Eight additional residue images are reserved and do not participate in selection or reported outcomes. \tabref{tab:data_splits} gives the complete per-repetition allocation and the role of each partition.

\begin{table}[!htbp]
\centering
\caption{Synthetic samples per end-to-end repetition. Held-out partitions do not influence weights, temperatures, candidate thresholds, or policy selection.}
\label{tab:data_splits}
\small
\begin{tabularx}{\linewidth}{>{\raggedright\arraybackslash}p{0.23\linewidth}r>{\raggedright\arraybackslash}p{0.31\linewidth}X}
\toprule
Partition & Images & Content & Use \\
\midrule
Training & 128 & nominal defects & model fitting \\
Validation calibration & 32 & nominal defects & temperature and candidate thresholds \\
Independent clean calibration & 16 & no defects & operating-policy selection \\
Nominal held-out & 48 & nominal defects & primary evaluation \\
Clean held-out & 24 & no defects & hallucination evaluation \\
Weak-defect held-out & 24 & attenuated nominal defects & weak-evidence stress \\
Residue held-out & 24 & unseen morphology & morphology-shift stress \\
Reserved residue & 8 & unseen morphology & excluded from selection and analysis \\
\bottomrule
\end{tabularx}
\end{table}

The defect composition is fixed at the schema level so methods see identical class counts within a seed. Bridge and gap defects alter connectivity or remove a local segment; missing-pattern defects remove a longer structure; particles add compact foreground; scratches create elongated responses. These morphologies stress different aspects of the thresholded output. Component recall is sensitive to whether a defect is touched at all, edge F1 is sensitive to boundary drift, and component IoU requires meaningful spatial agreement. \tabref{tab:defect_composition} reports the fixed family counts assigned to every partition.

\begin{table}[!htbp]
\centering
\caption{Defect composition per repetition. Clean-selection images are independent of validation calibration. Residue held-out and reserved residue are listed separately; reserved images are excluded from selection and analysis. The complete ten-run study contains ten times these counts.}
\label{tab:defect_composition}
\scriptsize
\setlength{\tabcolsep}{2.6pt}
\resizebox{\linewidth}{!}{%
\begin{tabular}{lrrrrrrrrr}
\toprule
Defect & Total & Train & Val.\ calib. & Clean select & Nominal test & Weak test & Clean test & Residue held-out & Residue reserved \\
\midrule
Bridge & 50 & 26 & 8 & 0 & 10 & 6 & 0 & 0 & 0 \\
Gap & 48 & 26 & 6 & 0 & 10 & 6 & 0 & 0 & 0 \\
Missing pattern & 42 & 24 & 6 & 0 & 8 & 4 & 0 & 0 & 0 \\
Particle & 46 & 26 & 6 & 0 & 10 & 4 & 0 & 0 & 0 \\
Scratch & 46 & 26 & 6 & 0 & 10 & 4 & 0 & 0 & 0 \\
No defect & 40 & 0 & 0 & 16 & 0 & 0 & 24 & 0 & 0 \\
Residue & 32 & 0 & 0 & 0 & 0 & 0 & 0 & 24 & 8 \\
\bottomrule
\end{tabular}}
\end{table}

\begin{figure}[!htbp]
  \centering
  \includegraphics[width=0.93\linewidth]{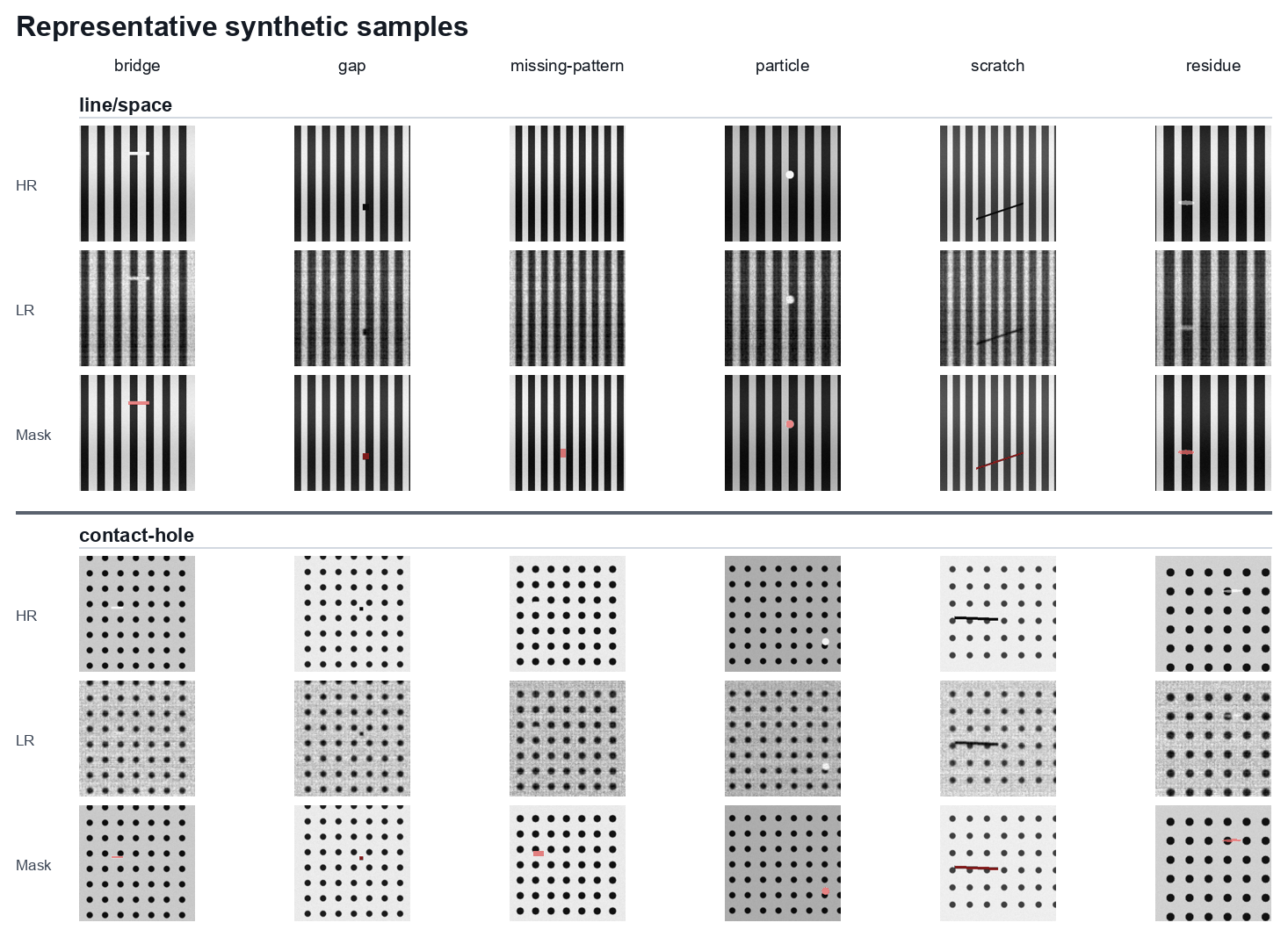}
  \caption{Representative line/space and contact-hole samples. Each generated item stores the high-resolution image, degraded low-resolution observation, defect mask, edge mask, clean-region mask, split label, defect family, and sample seed. The archive also stores the shared degradation configuration, so each stochastic observation can be regenerated deterministically. Weak and residue cases remain held out from model and policy selection.}
  \label{fig:dataset_examples}
\end{figure}

Split membership is generated from seed-fixed manifests rather than inferred from filenames during evaluation. Within each repetition, sample seeds are unique, exact high-resolution images are not duplicated, and fitting and held-out sample-seed sets are disjoint. Weak, residue, clean held-out, and external data are excluded from all model and operating-policy selection. Each seed changes both the generated data and neural initialization, so seed-level dispersion represents complete pipeline variation rather than training-only variation. \figref{fig:dataset_examples} shows representative line/space and contact-hole records together with their stored masks.

\subsection{Mask semantics}

The binary mask $M$ marks intended defect pixels. To avoid counting a small boundary displacement immediately as a clean hallucination, the clean-region mask excludes a three-pixel neighborhood of the defect:
\begin{equation}
C=1-\operatorname{dilate}(M,3).
\end{equation}
The edge mask is $E=\partial M$. Connected components are extracted from $M$ and $\widehat{M}_{\tau}$. A labeled component is counted by the permissive component-recall metric if any predicted component overlaps it. A stricter localization metric requires component IoU of at least 0.10. These definitions expose cases in which a detector touches many defects with only a few pixels but fails to localize their shapes.

Clean-only images use $C=1$ throughout the valid field. Their no-defect hallucination rate is the positive-pixel fraction, while the crop-call rate asks whether any pixel in a clean crop is called. The latter can be high even when the pixel fraction is small and therefore provides a review-oriented complement to FPR.

\subsection{Degradation model}

For high-resolution image $x$, the low-resolution observation is generated as
\begin{equation}
y=S_{\times1/2}\!\left[\operatorname{clip}\!\left(
a\{U_{k_d}(G_{\sigma_b}(x))-0.5\}+0.5+
\eta_{\mathrm{alias}}+\eta_{\mathrm{shot}}+\eta_{\mathrm{Poisson}}+
\eta_{\mathrm{Gaussian}}+\eta_{\mathrm{scan}}
\right)\right].
\label{eq:degradation}
\end{equation}
Here $G_{\sigma_b}$ is Gaussian blur, $U_{k_d}$ is a square uniform defocus filter with $k_d=\max(1,\operatorname{round}(3\sigma_d))$, $S$ is twofold downsampling, and $a$ is a contrast multiplier around mid-gray. The perturbations represent aliasing, shot variation, Poisson count variation, Gaussian electronic noise, and scan-line structure. Fourier optics, digital imaging, and sensor models motivate these factor families \cite{goodman2005fourier,gonzalez2008digital,janesick2001scientific}. Each archive stores one shared degradation configuration and a per-sample seed; together they reproduce the stochastic realization used to form the stored observation.

\begin{figure}[!htbp]
  \centering
  \includegraphics[width=\linewidth]{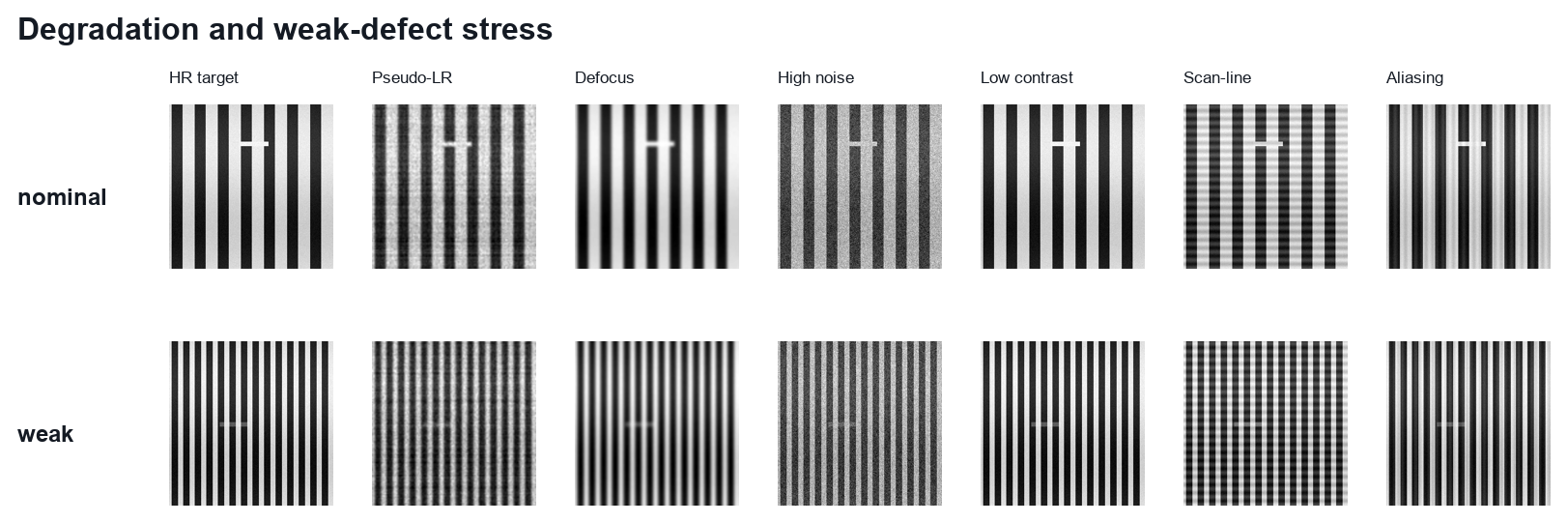}
  \caption{Illustrative effects of degradation and weak-defect settings. The panels explain the controlled factors; all quantitative results use the complete predeclared split and the full perturbation combination in Eq.~\eqref{eq:degradation}.}
  \label{fig:degradation_ablation}
\end{figure}

\tabref{tab:degradation} reports the single degradation configuration used by every headline repetition. These are controlled design settings rather than parameters estimated from a particular tool. This distinction is important: the controlled layer supports exact within-benchmark comparisons, while paired production measurements would be required to identify a scanner-specific degradation distribution. \figref{fig:degradation_ablation} varies selected factors only to illustrate their visible effects and weak-defect attenuation; it is not a second quantitative sweep.

\begin{table}[!htbp]
\centering
\caption{Headline degradation configuration. Stochastic realizations are determined by the per-sample seed.}
\label{tab:degradation}
\small
\begin{tabularx}{\linewidth}{>{\raggedright\arraybackslash}p{0.30\linewidth}p{0.24\linewidth}X}
\toprule
Factor & Headline setting & Interpretation \\
\midrule
Optical blur $\sigma_b$ & 1.2 & Gaussian point-spread variation \\
Uniform-defocus control $\sigma_d$ & 0.7 ($k_d=2$) & square-kernel focus variation \\
Downsampling & $2\times$ & low-resolution acquisition proxy \\
Shot-noise scale & 0.025 & signal-dependent variation \\
Gaussian-noise scale & 0.020 & electronic noise \\
Poisson peak & 80 & count variation \\
Scan-line amplitude & 0.030 & raster structure \\
Contrast multiplier $a$ & $\mathcal{U}(0.85,1.15)$ & per-image illumination/process drift \\
Aliasing amplitude & 0.020 & sampling artifact \\
\bottomrule
\end{tabularx}
\end{table}

\subsection{External SEM layer}

Carinthia-S contains 4,591 public SEM images and masks divided among six metadata classes \cite{carinthias2025}. Classes 1--5 contain defect-bearing images; class 6 is identified as no defect. The counts are strongly imbalanced: class 3 contains 4,008 images, class 4 contains 289, class 6 contains 227, class 1 contains 55, class 2 contains 8, and class 5 contains 4. This imbalance motivates two recall summaries. Defect-image macro recall averages per-image recall across all nonempty masks. Class-macro recall first averages within each of classes 1--5 and then gives each class equal weight.

Each SEM image is resized to $256\times256$ and bicubically downsampled to a $128\times128$ pseudo-low-resolution observation. The selected synthetic transformation or detector is applied, and the output is evaluated against the public mask. No external image changes weights, temperature, threshold, detector hyperparameter, or prior-fusion setting. Clean-region FPR excludes a three-pixel dilation of the mask, matching the synthetic definition. The no-defect crop-call rate is the fraction of the 227 class-6 images with at least one positive pixel.

The external layer answers a transfer question, not an acquisition-identification question. The pseudo-low-resolution image is generated from an SEM image rather than obtained through a paired optical measurement. Small classes make class-macro recall uncertain. These limitations are retained in the interpretation rather than hidden by pooling all external pixels into one dominant class.

%% file: sections/04_methods_protocol.tex
\section{Methods and Comparison Tracks}
\label{sec:methods}

The comparison is divided by the object being changed. Track~A changes only the image transformation before one common detector. Track~B contains models that predict the mask directly or jointly with a reconstruction. Both tracks use the same data partitions, operating-policy selection rule, clean-mask definition, thresholded metrics, and repeated-unit aggregation. Their outcomes are discussed together but are not interpreted as one architecture leaderboard.

\subsection{Track A: image transformations}

Every Track~A method maps the $128\times128$ observation to a clipped $256\times256$ image. The comparison set includes four interpolation rules, three classical restoration or sharpening rules, one deterministic NAF-inspired enhancement, and two trained compact reconstruction networks. \tabref{tab:transformations} defines each transformation. The comparison set distinguishes interpolation sensitivity, mild local enhancement, aggressive restoration, and reconstruction-only learning without allowing the downstream detector to change.

\begin{table}[!htbp]
\centering
\caption{Track~A image transformations. Every output is clipped to $[0,1]$, evaluated at $256\times256$ resolution, and scored by the identical detector in Eq.~\eqref{eq:common_detector}.}
\label{tab:transformations}
\small
\begin{tabularx}{\linewidth}{>{\raggedright\arraybackslash}p{0.24\linewidth}X}
\toprule
Method & Transformation \\
\midrule
Nearest, bilinear, bicubic, Lanczos & corresponding interpolation from $128\times128$ to $256\times256$ \\
Denoise--Lanczos & Gaussian smoothing with $\sigma=0.55$ at low resolution followed by Lanczos upsampling \\
Wiener & bicubic upsampling followed by Fourier-domain Wiener restoration with Gaussian PSF $\sigma=1.0$ and balance 0.03 \\
Sharpened & bicubic image plus 1.2 times its residual from Gaussian smoothing at $\sigma=1.0$ \\
NAF-style heuristic & bicubic detail residual at $\sigma=1.4$, weighted by normalized gradient magnitude with gain 0.9 \\
SR-lite & 24-channel, two-block residual reconstruction network with pixel-shuffle upsampling \\
Trained NAF-style SR & 24-channel, two-block gated NAF-style reconstruction network with pixel-shuffle upsampling \\
\bottomrule
\end{tabularx}
\end{table}

Bicubic interpolation is the single low-resolution reference. The low-resolution array cannot be compared pixelwise with a $256\times256$ mask until it is placed on the output grid; the benchmark uses bicubic interpolation for that placement. A second row labeled ``LR'' with the same bicubic pixels would not be another image intervention. This explicit equivalence prevents detector hyperparameters from creating a duplicate baseline.

SR-lite and the trained NAF-style model are fitted separately for each seed using the 128 paired training images. Both use AdamW with learning rate $10^{-3}$ and an $L_1$ reconstruction loss. Defect masks, clean masks, and edge masks do not enter either reconstruction loss. The deterministic NAF-style heuristic has no learned parameters. The reconstruction models therefore test the common assumption that better average image fidelity will also preserve sparse task evidence.

\subsection{One common downstream detector}

Every Track~A image $z$ is converted to a score by the same local-residual detector:
\begin{align}
r_z(u) &= |z(u)-G_{2.0}(z)(u)|,\
q_z(u) &= \operatorname{clip}\!\left(\frac{r_z(u)}{Q_{0.98}(r_z)+10^{-6}},0,1\right),\
p_z(u) &= G_{0.6}(q_z)(u).
\label{eq:common_detector}
\end{align}
Here $Q_{0.98}$ is the image-level 98th percentile. The $2.0$ residual scale, percentile normalization, final smoothing, temperature fitting, candidate targets, threshold selection, component extraction, and metrics are identical for every transformation. Only $z$ changes.

The detector is intentionally transparent. Periodic wafer structures create strong edges, while a defect perturbs the local pattern. Percentile normalization makes the score comparable across image transformations without using a label. The detector is not claimed to be optimal. Its role is experimental control: it converts image evidence into one fixed decision mechanism so differences among Track~A rows cannot be attributed to method-specific detector tuning.

\subsection{Track B: direct and joint predictors}

The primary direct detector uses a DeepLabV3-ResNet50 body without ImageNet initialization \cite{chen2017deeplabv3}. A learned $1\times1$ adapter maps the grayscale $128\times128$ input to three channels, and the output probability map is bilinearly resized by two. Training uses AdamW at learning rate $10^{-3}$, batch size 4, positive-pixel weight 64, clean-region loss weight 1, and edge-loss weight 0.1. Only the training partition updates the weights. The selected model state is determined without held-out outcomes, and evaluation refits the scalar probability temperature on the validation calibration partition.

DeepLabV3 differs substantially in capacity and objective from the compact reconstruction models. This is deliberate. Its result tests whether the low-resolution observation contains task evidence that mask supervision can recover. It does not establish that DeepLabV3 is a better SR architecture or that capacity alone explains the difference. A capacity-matched task-aware SR experiment is identified as a next step in \secref{sec:limitations}.

\begin{figure}[!htbp]
  \centering
  \includegraphics[width=\linewidth]{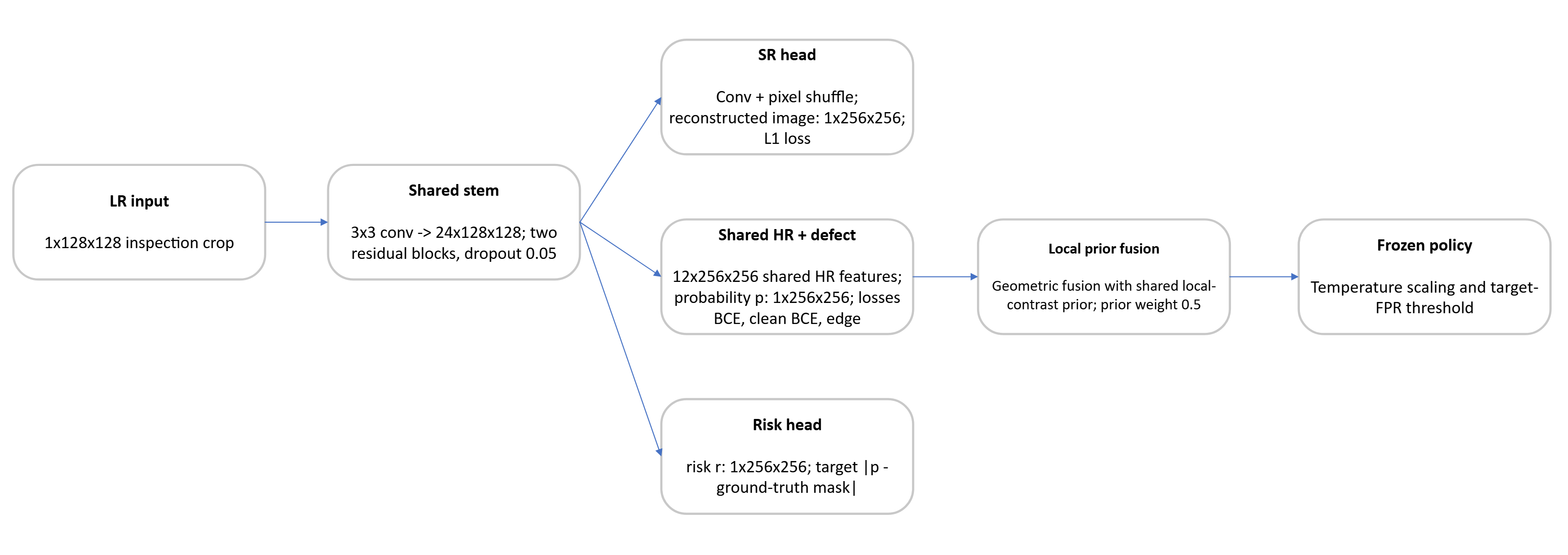}
  \caption{DPU-WaferSR used as the illustrative joint participant. The compact trunk feeds reconstruction, defect, and risk heads. Prior fusion, scalar temperature fitting, and operating-point selection are applied after inference and are evaluated explicitly rather than treated as part of an SR-only comparison.}
  \label{fig:architecture}
\end{figure}

DPU-WaferSR is an illustrative joint participant. It has a 24-channel stem, two residual blocks with dropout 0.05, a pixel-shuffle reconstruction head, a shared high-resolution feature block, and separate defect and risk heads. The model predicts $(\widehat{x},p,r)$ from $y$. Four dropout samples are averaged at evaluation, and the defect probability is geometrically fused with a local residual prior at weight 0.5 before temperature fitting and threshold selection. Its purpose here is not to establish a new architecture family. It supplies a policy whose clean-calibration and held-out behavior can be followed under the same no-test-feedback rule. \figref{fig:architecture} summarizes the shared trunk, three prediction heads, and post-inference operating-policy stages.

The joint training objective is
\begin{equation}
\mathcal{L}=\mathcal{L}_{\mathrm{recon}}+
\mathcal{L}_{\mathrm{defect}}+2\mathcal{L}_{\mathrm{clean}}+
0.2\mathcal{L}_{\mathrm{edge}}+0.05\mathcal{L}_{\mathrm{risk}},
\label{eq:dpu_loss}
\end{equation}
where
\begin{align}
\mathcal{L}_{\mathrm{recon}} &= \|\widehat{x}-x\|_1,\\
\mathcal{L}_{\mathrm{defect}} &= \operatorname{BCE}_{w_+=64}(p,M),\\
\mathcal{L}_{\mathrm{clean}} &= \operatorname{BCE}(p(C),0),\\
\mathcal{L}_{\mathrm{edge}} &= 1-\operatorname{Dice}(p,E),\\
\mathcal{L}_{\mathrm{risk}} &= \|r-|\operatorname{stopgrad}(p)-M|\|_1.
\end{align}
The model is trained for five epochs with AdamW, learning rate $10^{-3}$, and batch size 4. The reconstruction term preserves image signal, the weighted defect term addresses sparse positives, the clean term penalizes responses away from defect neighborhoods, the edge term encourages boundary agreement, and the detached risk target supports risk-coverage analysis without feeding its error back into the defect target. \tabref{tab:model_profiles} reports the parameter and multiply--accumulate scales of all learned participants.

\begin{table}[!htbp]
\centering
\caption{Neural model profiles for a $128\times128$ input and $256\times256$ output. Multiply--accumulate counts are approximate convolution counts from forward hooks.}
\label{tab:model_profiles}
\small
\begin{tabular}{lrrl}
\toprule
Model & Parameters & Approximate MACs & Supervision \\
\midrule
SR-lite & 42,121 & 697,171,968 & paired image $L_1$ \\
Trained NAF-style SR & 25,993 & 428,212,224 & paired image $L_1$ \\
DeepLabV3 detector & 39,633,735 & 10,222,944,256 & defect, clean, edge masks \\
DPU-WaferSR & 44,883 & 875,692,032 & reconstruction, defect, clean, edge, risk \\
\bottomrule
\end{tabular}
\end{table}

\section{Operating-Point Protocol and Metrics}
\label{sec:protocol}

The operating-point procedure is the same for every reported method and seed. It separates probability fitting, independent clean selection, and held-out evaluation. This separation is the mechanism that makes a failed false-positive constraint scientifically visible.

\subsection{Three-stage policy selection}

The candidate target set is
\begin{equation}
\mathcal{A}=\{3,2,1.5,1,0.75,0.5,0.3\}\times10^{-4},
\end{equation}
ordered from most to least permissive. For each method and seed:

\begin{enumerate}
  \item Fit a scalar temperature on the 32 validation calibration images by minimizing pixelwise binary cross-entropy.
  \item On those validation images, obtain one threshold for each target in $\mathcal{A}$.
  \item Evaluate the candidates in order on the 16 independent clean calibration images. Select the first threshold whose mean clean-region FPR is at most $4.5\times10^{-4}$.
  \item If no candidate satisfies the criterion, retain the most conservative candidate and record a selection failure.
  \item Without further fitting, evaluate nominal, clean, weak, and residue held-out partitions once. When external transfer is reported, apply the same transformation or model, temperature, and threshold to Carinthia-S.
\end{enumerate}

\begin{figure}[!htbp]
  \centering
  \includegraphics[width=0.90\linewidth]{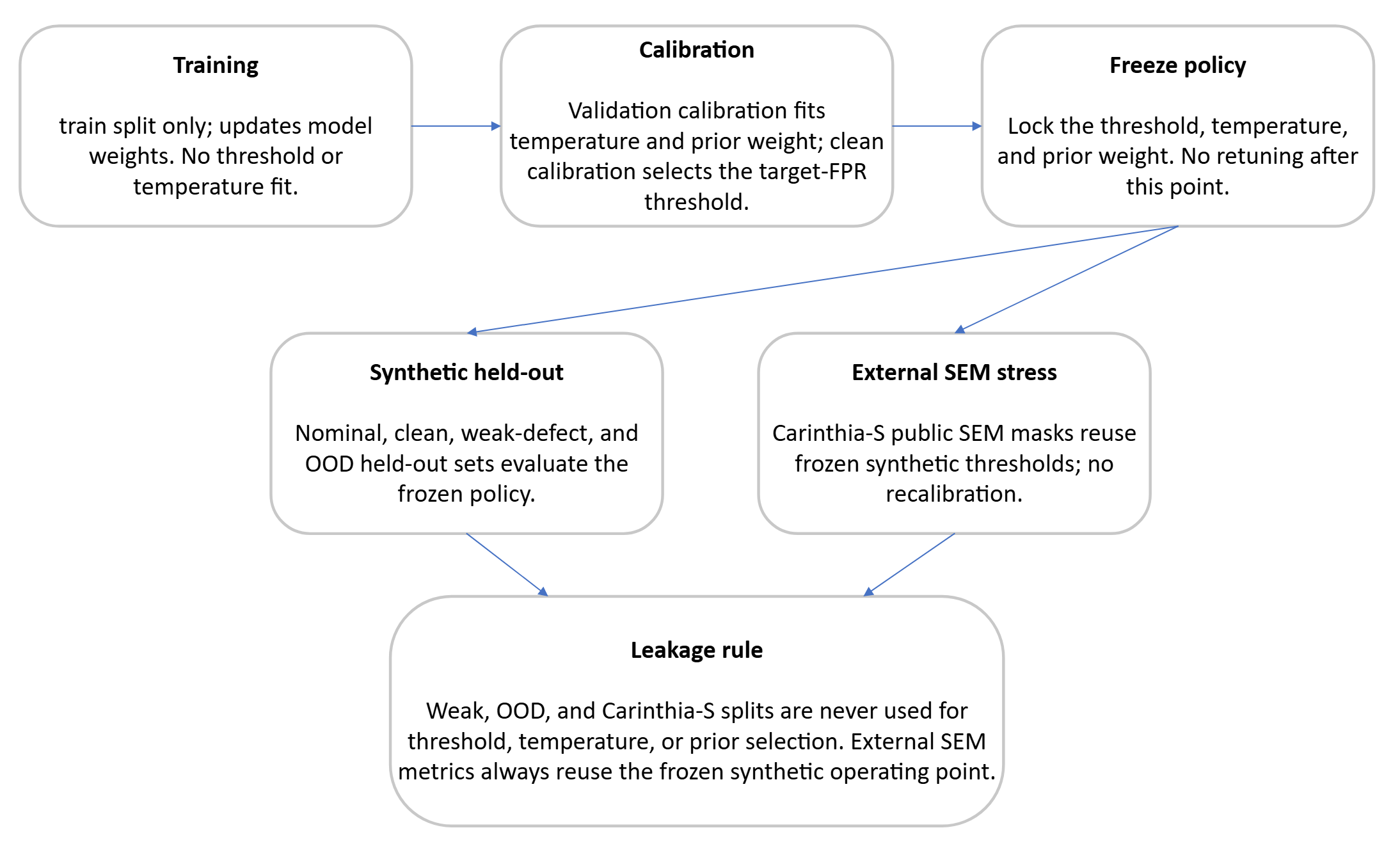}
  \caption{Separation of fitting, independent clean selection, and held-out evaluation. Training data update weights; validation data fit temperature and candidate thresholds; clean calibration data choose the most permissive acceptable candidate; all reported held-out and external outcomes are measured after the policy is fixed.}
  \label{fig:calibration_protocol}
\end{figure}

No held-out recall, FPR, weak metric, residue metric, or external metric participates in this rule. The clean selection partition contains no defects and therefore cannot optimize recall. It can only reject candidate policies whose clean activation is too high. \figref{fig:calibration_protocol} makes the separation among fitting, clean selection, and held-out evaluation explicit.

This design is stricter than choosing the best test-feasible candidate after observing the test set. It also produces an interpretable failure. If a policy satisfies the independent clean partition and exceeds the nominal held-out limit, the result measures a change in background structure or score distribution. Choosing a stricter target after seeing that outcome would remove the evidence of transfer failure.

\subsection{Primary pixel metrics}

For each image with a nonempty defect mask, defect-pixel recall is
\begin{equation}
\mathrm{Recall}_{\tau}=\frac{|\widehat{M}_{\tau}\cap M|}{|M|+\epsilon}.
\end{equation}
Clean-region false-positive rate is
\begin{equation}
\fpr_{\tau}=\frac{|\widehat{M}_{\tau}\cap C|}{|C|+\epsilon}.
\end{equation}
Metrics are averaged over images within a seed before seed-level aggregation. This image-first convention prevents a large mask or crop from determining the run average. Pixel precision and mask IoU are
\begin{align}
\mathrm{Precision}_{\tau} &= \frac{|\widehat{M}_{\tau}\cap M|}{|\widehat{M}_{\tau}|+\epsilon},\\
\mathrm{IoU}_{\tau} &= \frac{|\widehat{M}_{\tau}\cap M|}{|\widehat{M}_{\tau}\cup M|+\epsilon}.
\end{align}

The no-defect hallucination rate $\nhr$ applies the FPR numerator and denominator to clean-only images. The crop-call rate is the fraction of clean images for which $|\widehat{M}_{\tau}|>0$. These measures answer different questions: NHR measures how many clean pixels activate, whereas crop-call rate measures how often an image would be routed for review by at least one call.

\subsection{Component, boundary, and reconstruction metrics}

Let $\mathcal{K}(M)$ denote the connected components of the labeled mask. Permissive component recall is
\begin{equation}
\mathrm{CompRecall}_{\tau}=\frac{1}{|\mathcal{K}(M)|}
\sum_{K\in\mathcal{K}(M)}
\mathbb{1}[|\widehat{M}_{\tau}\cap K|>0].
\end{equation}
The stricter component metric requires a matched predicted component to reach IoU $\geq0.10$. Edge F1 compares the boundary of $\widehat{M}_{\tau}$ with $E$. Together, these outcomes distinguish a sparse touch from meaningful localization.

PSNR and SSIM are computed only for image-producing methods. They are descriptive reconstruction outcomes rather than surrogates for inspection utility. A model can improve these scores by reconstructing the common periodic background accurately while attenuating a sparse defect. The paired analysis in \secref{sec:track_a_results} tests this relationship directly.

\subsection{Weak, morphology-shift, and external outcomes}

Weak-defect recall uses the same selected policy on attenuated versions of the nominal defect families. Residue recall uses the held-out morphology. Neither split changes the policy. External defect-image macro recall averages each nonempty Carinthia-S mask equally. External class-macro recall gives equal weight to metadata classes 1--5 after averaging within class. External clean-region FPR follows the same three-pixel exclusion rule, and no-defect crop-call rate uses class 6.

These outcomes are reported separately rather than combined into one composite score. Nominal feasibility, weak sensitivity, new-morphology sensitivity, and external background transfer represent distinct failure modes. A weighted average would obscure which regime caused a method to fail.

\subsection{Repeated unit and paired analysis}

Each of the ten seeds defines a complete repetition: generated data, learned weights, fitted temperature, candidate thresholds, independent clean selection, and held-out evaluation. Within a repetition, image-level metrics are averaged. Across repetitions, tables report the arithmetic mean and population standard deviation of the ten seed summaries. The population denominator describes dispersion of this predeclared run set and is stated to avoid ambiguity.

Method differences are paired by seed. For method $m$ and the bicubic reference $b$, the recall difference is $\Delta_s=R_{m,s}-R_{b,s}$. We report the mean and standard deviation of $\Delta_s$ and the number of positive or negative directions. This effect-size-first reporting avoids a large family of unadjusted hypothesis tests and does not treat pixels as independent observations. Pooled clean-pixel counts can be checked for arithmetic consistency, but seed-level variation remains the inferential unit.

\subsection{Implementation}

The implementation uses PyTorch, NumPy, Pillow, and deterministic seed manifests. Neural training and external inference run on an NVIDIA GeForce RTX~5090. CUDA timing is synchronized when runtime is collected; runtime is treated as an implementation measurement rather than a production-throughput claim. All primary summary values are first computed within each complete seed-level repetition and then aggregated across the ten repetitions.

%% file: sections/05_results.tex
\section{Results}
\label{sec:results}

The results answer four questions. First, does reconstruction fidelity predict preserved defect evidence when the downstream detector is held fixed? Second, does a policy that is feasible on calibration data remain feasible on held-out data? Third, do weak defects and a held-out morphology remain detectable under the same frozen policies? Fourth, what transfers when those policies are applied to public SEM masks without retraining or recalibration? The answer is consistent across the synthetic and external layers: image similarity, calibration success, weak-defect sensitivity, and held-out inspection safety are different properties. All synthetic values are mean $\pm$ SD over the ten predeclared repetitions unless otherwise stated.

\subsection{Reconstruction fidelity does not predict preserved defect evidence}
\label{sec:track_a_results}

Track~A isolates the effect of the image transformation. Each method produces a $256\times256$ image from the same $128\times128$ observation, and every output is scored by the same local residual detector with $\sigma=2.0$. Under that control, the bicubic reference is not a second detector setting or a special low-resolution row; it is the single upsampled representation of the low-resolution observation.

\begin{figure}[!htbp]
  \centering
  \includegraphics[width=\linewidth]{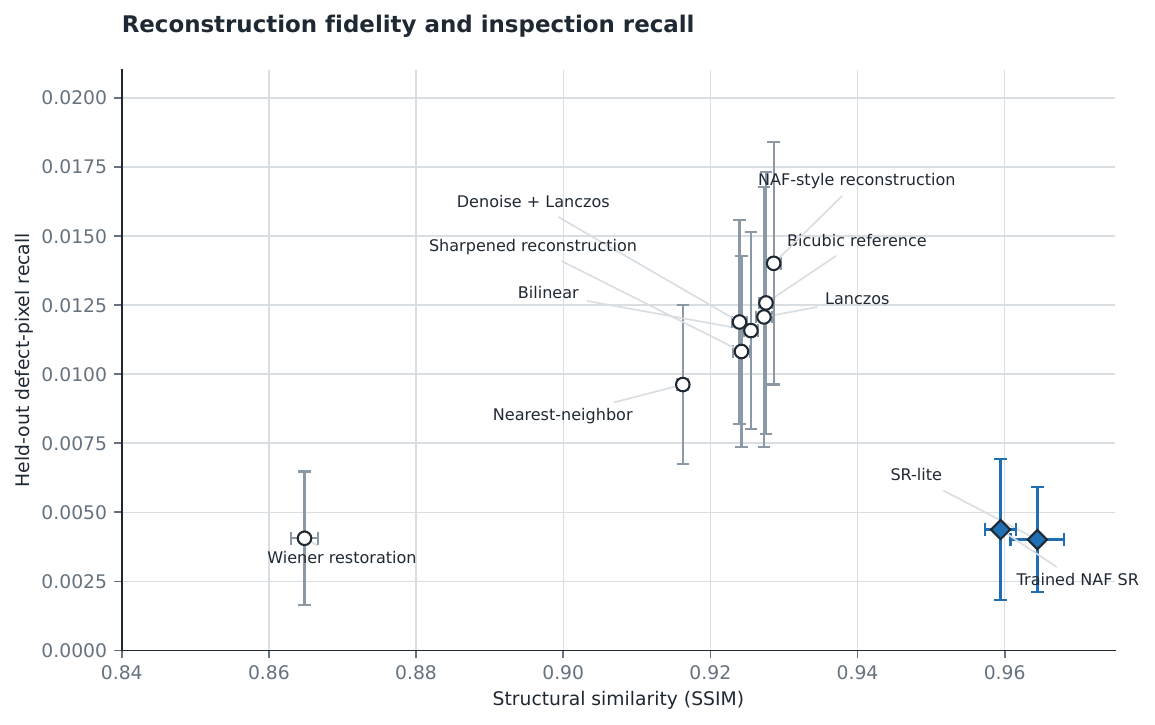}
  \caption{Reconstruction fidelity versus defect-evidence preservation. Markers show reconstruction-method means, and error bars show seed-level standard deviations. The learned reconstruction methods attain high structural similarity but low defect recall under the same downstream detector. DeepLabV3 belongs to the separate Track~B direct-detector comparison and is therefore not plotted in this reconstruction-only panel.}
  \label{fig:reconstruction_relationship}
\end{figure}

\figref{fig:reconstruction_relationship} shows the central relationship. The learned reconstruction models improve SSIM substantially, but their defect recall decreases. NAF-trained SR reaches SSIM $0.9594\pm0.0021$, and SR-lite reaches $0.9644\pm0.0037$, yet both recover only about one third of bicubic's already-low pixel recall. The result is not caused by a stricter false-call budget for the learned models: all Track~A methods satisfy the held-out feasibility criterion in all ten repetitions. \tabref{tab:primary_synthetic_results} gives the full Track~A comparison and the separate DeepLabV3 reference.

\begin{table}[!htbp]
\centering
\caption{Primary held-out synthetic results under the unified protocol. FPR is reported in units of $10^{-4}$. Track~A rows use one common local residual detector. The DeepLabV3 row is a Track~B direct-detector reference; SSIM is omitted because DeepLabV3 is not an image-producing method.}
\label{tab:primary_synthetic_results}
\scriptsize
\setlength{\tabcolsep}{3.2pt}
\resizebox{\linewidth}{!}{%
\begin{tabular}{llrrrrrr}
\toprule
Method & Track & Recall & FPR & SSIM & Precision & Component recall & Feasible runs \\
\midrule
Nearest & A & $0.0096\pm0.0029$ & $1.2801\pm0.3070$ & $0.9163\pm0.0008$ & $0.202\pm0.023$ & $0.285\pm0.028$ & 10/10 \\
Bilinear & A & $0.0116\pm0.0036$ & $0.7498\pm0.2339$ & $0.9255\pm0.0009$ & $0.218\pm0.031$ & $0.298\pm0.035$ & 10/10 \\
Bicubic & A & $0.0126\pm0.0048$ & $0.6037\pm0.2388$ & $0.9276\pm0.0010$ & $0.232\pm0.024$ & $0.325\pm0.036$ & 10/10 \\
Lanczos & A & $0.0121\pm0.0047$ & $0.7623\pm0.2500$ & $0.9273\pm0.0010$ & $0.232\pm0.038$ & $0.327\pm0.041$ & 10/10 \\
Denoise+Lanczos & A & $0.0119\pm0.0037$ & $1.2947\pm0.3848$ & $0.9240\pm0.0010$ & $0.203\pm0.028$ & $0.296\pm0.042$ & 10/10 \\
Wiener & A & $0.0041\pm0.0024$ & $0.8425\pm0.3816$ & $0.8648\pm0.0019$ & $0.131\pm0.029$ & $0.196\pm0.056$ & 10/10 \\
Sharpened & A & $0.0108\pm0.0035$ & $0.3556\pm0.1214$ & $0.9242\pm0.0011$ & $0.237\pm0.023$ & $0.306\pm0.026$ & 10/10 \\
NAF heuristic & A & $0.0140\pm0.0044$ & $0.3927\pm0.1416$ & $0.9286\pm0.0010$ & $0.255\pm0.024$ & $0.335\pm0.020$ & 10/10 \\
SR-lite & A & $0.0040\pm0.0019$ & $0.1943\pm0.1574$ & $0.9644\pm0.0037$ & $0.132\pm0.022$ & $0.150\pm0.026$ & 10/10 \\
NAF trained & A & $0.0044\pm0.0025$ & $0.4265\pm0.6610$ & $0.9594\pm0.0021$ & $0.139\pm0.036$ & $0.167\pm0.046$ & 10/10 \\
DeepLabV3 & B & $0.1984\pm0.0385$ & $1.7431\pm0.8433$ & --- & $0.256\pm0.048$ & $0.413\pm0.076$ & 10/10 \\
\bottomrule
\end{tabular}}
\end{table}

\tabref{tab:paired_deltas} reports the paired view against bicubic. The deterministic NAF heuristic improves mean recall slightly in seven of ten repetitions, but the effect is small relative to the remaining recall gap. In contrast, the two trained reconstruction models lose recall in every repetition. The direct DeepLabV3 detector gains recall in every repetition, which indicates that the low-resolution observation still contains usable mask evidence when the prediction problem is trained directly. That result belongs to Track~B; it does not rescue the claim that higher-fidelity reconstruction preserves evidence under a fixed detector.

\begin{table}[!htbp]
\centering
\caption{Paired recall changes relative to bicubic under the same repeated seeds. Values are seed-level paired differences on nominal held-out data.}
\label{tab:paired_deltas}
\small
\begin{tabular}{lrrr}
\toprule
Comparison & Recall delta & Positive seeds & Negative seeds \\
\midrule
NAF heuristic $-$ bicubic & $+0.0014\pm0.0027$ & 7/10 & 3/10 \\
NAF trained $-$ bicubic & $-0.0082\pm0.0045$ & 0/10 & 10/10 \\
SR-lite $-$ bicubic & $-0.0086\pm0.0041$ & 0/10 & 10/10 \\
DeepLabV3 $-$ bicubic & $+0.1859\pm0.0383$ & 10/10 & 0/10 \\
\bottomrule
\end{tabular}
\end{table}

Component and boundary measures tell the same story. Bicubic reaches component recall $0.325\pm0.036$, while SR-lite and NAF-trained SR fall to $0.150\pm0.026$ and $0.167\pm0.046$. Their IoU-at-0.10 component hits remain near zero, showing that most detected pixels do not form useful localized components. DeepLabV3 improves component recall and IoU-at-0.10, but weak-defect recall remains very low even for that stronger direct detector. \tabref{tab:secondary_metrics} reports the corresponding precision, component, boundary, and clean-control measures.

\begin{table}[!htbp]
\centering
\caption{Secondary held-out metrics for representative methods. Clean hallucination rate is measured on clean-only controls and reported in units of $10^{-4}$.}
\label{tab:secondary_metrics}
\scriptsize
\setlength{\tabcolsep}{4pt}
\begin{tabular}{lrrrrr}
\toprule
Method & Precision & Component recall & Component IoU$\geq0.10$ & Edge F1 & Clean hallucination \\
\midrule
Bicubic & $0.232\pm0.024$ & $0.325\pm0.036$ & $0.025\pm0.018$ & $0.015\pm0.007$ & $1.668\pm0.368$ \\
SR-lite & $0.132\pm0.022$ & $0.150\pm0.026$ & $0.004\pm0.008$ & $0.008\pm0.004$ & $2.022\pm0.454$ \\
NAF trained & $0.139\pm0.036$ & $0.167\pm0.046$ & $0.006\pm0.010$ & $0.006\pm0.004$ & $2.109\pm0.759$ \\
DeepLabV3 & $0.256\pm0.048$ & $0.413\pm0.076$ & $0.356\pm0.070$ & $0.162\pm0.031$ & $0.000\pm0.000$ \\
\bottomrule
\end{tabular}
\end{table}

\begin{figure}[!htbp]
  \centering
  \includegraphics[width=\linewidth]{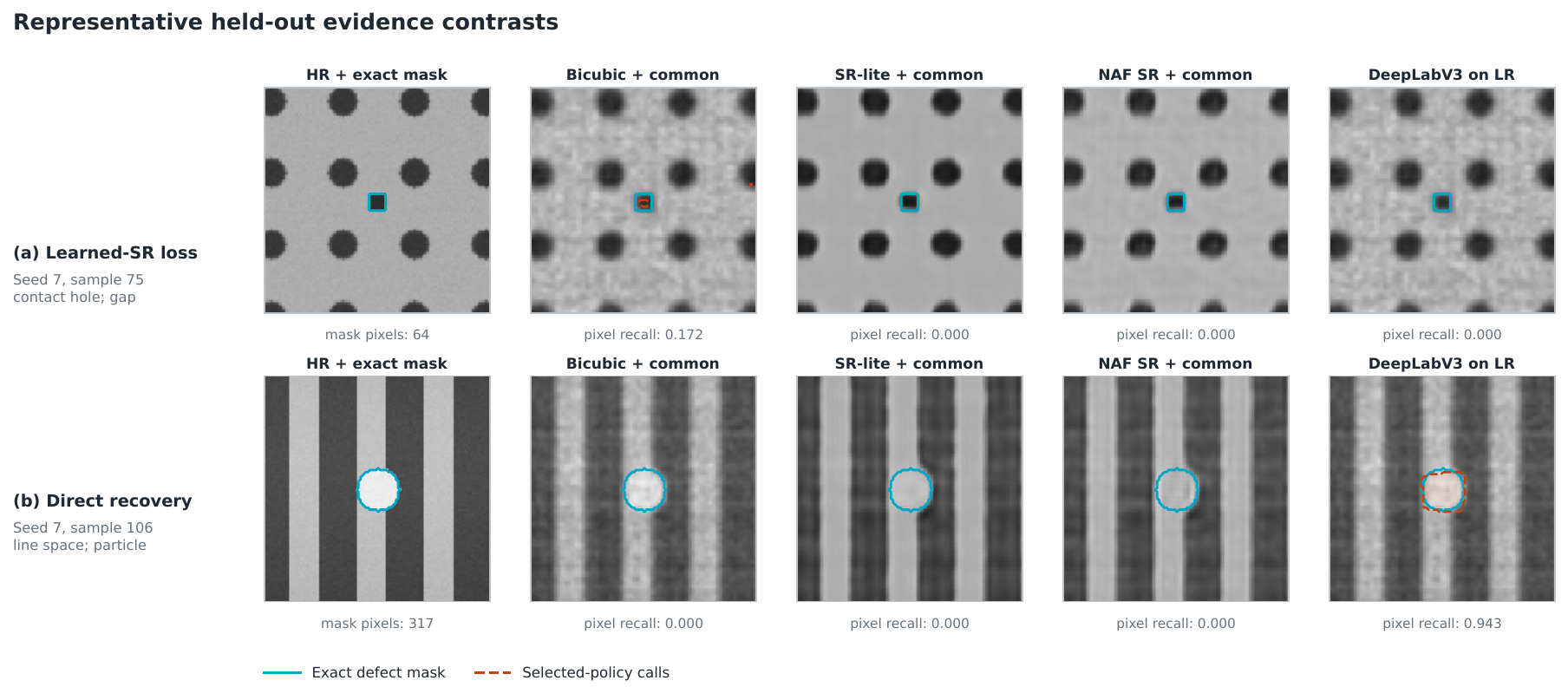}
  \caption{Representative held-out evidence contrasts under preselected seed-7 policies. All panels in a row show the same defect-centered crop. Solid cyan contours mark the exact mask; dashed orange contours and translucent fill mark thresholded calls. Row (a) is the nominal test image with the largest bicubic-over-learned-reconstruction recall contrast, and row (b) is the distinct image with the largest DeepLabV3-over-bicubic contrast. Per-image recalls are printed below the panels; aggregate conclusions use all ten repetitions rather than these examples.}
  \label{fig:qualitative_evidence}
\end{figure}

\figref{fig:qualitative_evidence} makes the paired contrast visible in individual held-out images. Both rows come from the nominal test split of seed 7 and use policies fixed before testing. Row (a) maximizes the per-image recall advantage of bicubic over the better of the two learned reconstructions; row (b) maximizes the per-image recall advantage of DeepLabV3 over bicubic, with a distinct image enforced. The learned reconstructions remain visually plausible even when the common detector loses mask evidence, whereas the direct detector can recover a larger portion of the same annotated region. These examples illustrate the measured mechanism but do not enter the ten-run summary estimates.

\subsection{Clean calibration does not guarantee held-out false-call control}
\label{sec:operating_transfer_results}

The second result moves from Track~A reconstructions to Track~B predictors under the same three-stage operating-point rule. DeepLabV3 is a direct mask predictor trained on the low-resolution observation. DPU-WaferSR is an illustrative joint reconstruction/detection model that produces a reconstructed image together with defect and risk scores. These models are not compared as alternative SR transforms for a fixed detector; they test whether a task-trained policy that is feasible on independent clean calibration remains feasible on held-out images.

The unified selection rule fits temperature and candidate thresholds on validation data, chooses the most permissive feasible candidate on independent clean images, and evaluates held-out images once. This design prevents test-set threshold selection and makes a failed held-out policy a meaningful outcome rather than a tuning inconvenience.

\begin{figure}[!htbp]
  \centering
  \includegraphics[width=\linewidth]{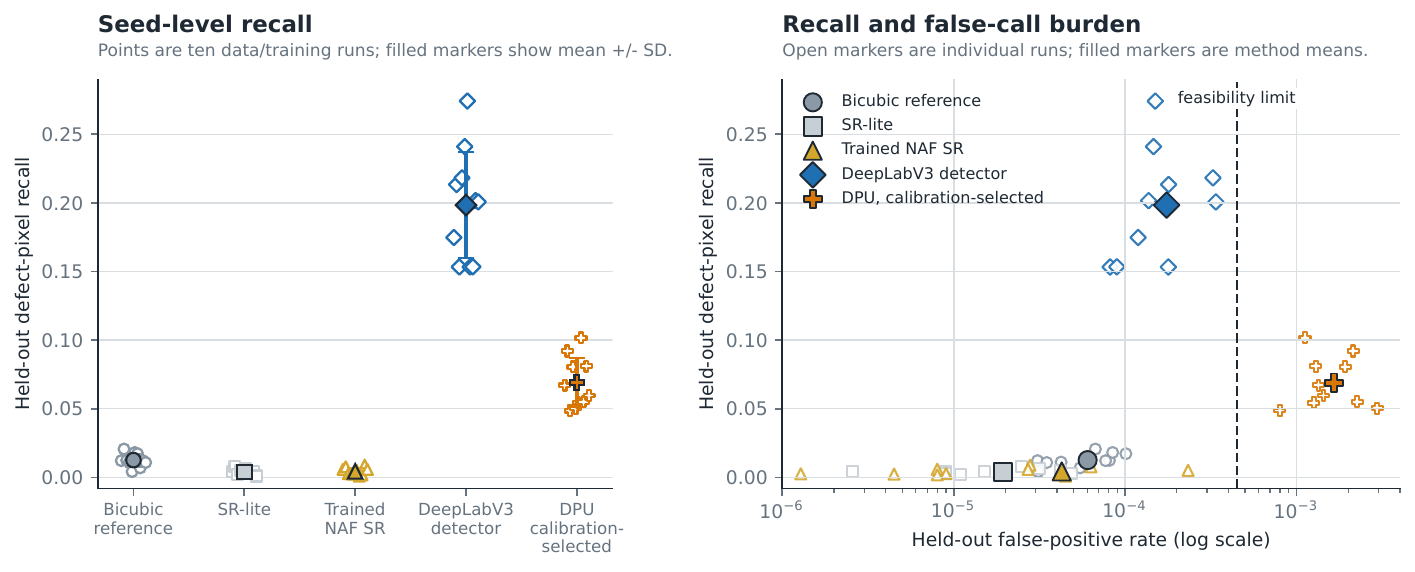}
  \caption{Operating-point transfer under the common three-stage rule. Open markers show seed policies and filled markers show method means. The dashed line marks the held-out feasibility limit of $4.5\times10^{-4}$ FPR. The DPU policy is selected without test feedback and then fails held-out false-call control.}
  \label{fig:unified_protocol}
\end{figure}

\figref{fig:unified_protocol} compares the selected policies. Bicubic remains the Track~A reference: quiet and feasible, but low recall. DeepLabV3 achieves much higher nominal recall while remaining feasible in all ten held-out repetitions. DPU-WaferSR passes independent clean calibration in all ten repetitions, but its held-out FPR rises to $0.001650\pm0.000615$, so none of the ten held-out repetitions meets the $4.5\times10^{-4}$ feasibility limit. \tabref{tab:operating_transfer} separates selection feasibility, held-out feasibility, weak recall, and clean hallucination for the three representative policies.

\begin{table}[!htbp]
\centering
\caption{Operating-point transfer for representative methods. FPR and clean hallucination rate are in units of $10^{-4}$; weak recall is in units of $10^{-3}$. Selection feasibility is measured on independent clean calibration images, and held-out feasibility is measured on nominal test images.}
\label{tab:operating_transfer}
\scriptsize
\setlength{\tabcolsep}{4pt}
\resizebox{\linewidth}{!}{%
\begin{tabular}{llrrrrrr}
\toprule
Method & Track / role & Recall & Test FPR & Weak recall & Clean hallucination & Selection feasible & Held-out feasible \\
\midrule
Bicubic + common detector & A / reconstruction reference & $0.0126\pm0.0048$ & $0.604\pm0.239$ & $0.257\pm0.568$ & $1.668\pm0.368$ & 10/10 & 10/10 \\
DeepLabV3 direct detector & B / direct prediction & $0.1984\pm0.0385$ & $1.743\pm0.843$ & $0.612\pm1.351$ & $0.000\pm0.000$ & 10/10 & 10/10 \\
DPU-WaferSR joint model & B / joint reconstruction+detection & $0.0690\pm0.0177$ & $16.496\pm6.146$ & $3.253\pm2.959$ & $4.55\pm1.99$ & 10/10 & 0/10 \\
\bottomrule
\end{tabular}}
\end{table}

The DPU result is important precisely because it is unfavorable. The selected policy is not a post-hoc failure case and not a threshold accidentally chosen on test labels. It is the outcome of a predeclared rule applied to independent clean calibration data. The held-out images include defect-adjacent structure and a different mixture of patterned clean regions, and those differences are enough to break false-call control. For inspection, a threshold is therefore a transferable policy, not just a scalar.

\subsection{Weak defects and held-out morphology remain unsolved}
\label{sec:weak_results}

The benchmark is not saturated by the nominal result. Feasible methods recover almost none of the weak-defect pixels. Bicubic weak-defect recall is $0.000257\pm0.000568$, NAF-trained SR is $0.000045\pm0.000070$, SR-lite is $0.000027\pm0.000081$, and DeepLabV3 is $0.000612\pm0.001351$. DPU-WaferSR has higher weak-defect recall, $0.003253\pm0.002959$, but at an infeasible held-out false-call burden.

The held-out residue morphology is also difficult. Track~A residue recall is effectively zero under the selected common-detector thresholds, and DeepLabV3 reaches only $0.003797\pm0.007147$. This combination is the main scientific pressure point for future methods: nominal recall can improve while weak evidence and morphology transfer remain almost absent. \tabref{tab:weak_residue} places nominal, weak, and residue recall on one scale.

\begin{table}[!htbp]
\centering
\caption{Weak-defect and held-out-morphology behavior. Weak recall is in units of $10^{-3}$; residue recall is measured on the held-out morphology split. Residue recall is omitted for DPU-WaferSR because the selected joint policy is already held-out infeasible on the primary FPR criterion.}
\label{tab:weak_residue}
\small
\begin{tabular}{lrrr}
\toprule
Method & Nominal recall & Weak recall & Residue recall \\
\midrule
Bicubic & $0.0126\pm0.0048$ & $0.257\pm0.568$ & $0.0000\pm0.0000$ \\
SR-lite & $0.0040\pm0.0019$ & $0.027\pm0.081$ & $0.0000\pm0.0000$ \\
NAF trained & $0.0044\pm0.0025$ & $0.045\pm0.070$ & $0.0000\pm0.0000$ \\
DeepLabV3 & $0.1984\pm0.0385$ & $0.612\pm1.351$ & $0.0038\pm0.0071$ \\
DPU-WaferSR & $0.0690\pm0.0177$ & $3.253\pm2.959$ & --- \\
\bottomrule
\end{tabular}
\end{table}

\subsection{External SEM masks expose method-dependent transfer shifts}
\label{sec:external_results}

The public Carinthia-S stress layer contains 4,591 SEM masks with class counts 55, 8, 4,008, 289, 4, and 227 across the six released source classes. No Carinthia-S image or mask is used to train a model, fit a temperature, choose a threshold, or tune a prior. Each synthetic policy is applied unchanged to the external images after pseudo-low-resolution formation.

\begin{figure}[!htbp]
  \centering
  \includegraphics[width=\linewidth]{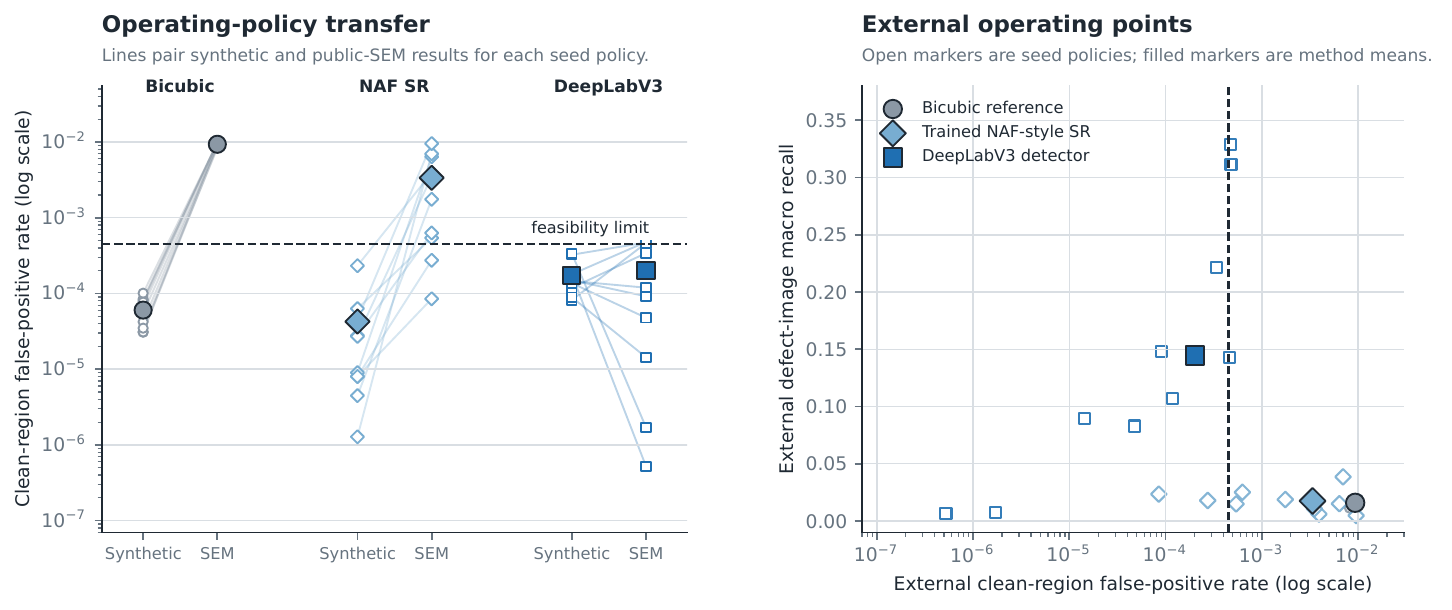}
  \caption{External transfer to 4,591 Carinthia-S masks under unchanged synthetic policies. Left: each line pairs one seed policy's synthetic and external clean-region FPR. Right: external defect-image macro recall versus external FPR. No external image is used for training, calibration, or operating-point selection.}
  \label{fig:external_transfer}
\end{figure}

\figref{fig:external_transfer} and \tabref{tab:external_transfer} show that external texture changes the false-call behavior sharply. The bicubic/common-detector policy has external clean-region FPR $0.00935\pm0.00057$, and every no-defect SEM crop receives at least one called pixel. NAF-trained SR lowers mean external FPR to $0.00337\pm0.00316$, but $95.99\%\pm6.21\%$ of no-defect crops still receive a call. DeepLabV3 transfers more quietly, with external FPR $0.000203\pm0.000199$ and no-defect crop-call rate $0.088\%\pm0.264\%$, but only seven of ten seed policies satisfy the same $4.5\times10^{-4}$ feasibility limit externally.

\begin{table}[!htbp]
\centering
\caption{Carinthia-S transfer under unchanged synthetic policies. Clean-region FPR is in units of $10^{-4}$; no-defect crop calls report the percentage of class-6 images containing at least one predicted pixel.}
\label{tab:external_transfer}
\small
\resizebox{\linewidth}{!}{%
\begin{tabular}{lrrrrr}
\toprule
Method & Image-macro recall & Class-macro recall & Clean FPR & No-defect calls & Feasible policies \\
\midrule
Bicubic + common detector & $0.0159\pm0.0024$ & $0.00629\pm0.00084$ & $93.48\pm5.67$ & $100.0\pm0.0\%$ & 0/10 \\
NAF trained + common detector & $0.0175\pm0.0094$ & $0.00617\pm0.00260$ & $33.70\pm31.55$ & $95.99\pm6.21\%$ & 2/10 \\
DeepLabV3 direct detector & $0.1446\pm0.1067$ & $0.0820\pm0.0597$ & $2.031\pm1.990$ & $0.088\pm0.264\%$ & 7/10 \\
\bottomrule
\end{tabular}}
\end{table}

The external ranking should be interpreted as stress evidence rather than as a production SEM benchmark. The SEM texture, pseudo-low-resolution formation, and class imbalance differ from the synthetic optical-style layer. Nevertheless, the stress test fulfills its purpose: a policy that is quiet on synthetic held-out images can activate broadly on real SEM backgrounds, and the magnitude of that activation depends on whether the score comes from a local reconstruction detector or a direct mask predictor.

\subsection{Result synthesis}
\label{sec:result_synthesis}

The four results support one story. The common-detector reconstruction track shows that high image similarity can remove the residual evidence a low-FPR detector needs. The operating-point experiment shows that clean calibration does not ensure held-out false-call transfer. The weak and morphology-shift splits show that nominal performance does not imply robustness to faint or unseen defects. The external SEM layer shows that the same policy can change false-call behavior under real texture even when no external tuning is allowed. A high-quality inspection SR study must therefore report the decision policy and its transfer, not only the reconstructed image.

%% file: sections/06_discussion.tex
\section{Discussion}
\label{sec:discussion}

\subsection{What the controlled comparison changes}
\label{sec:discussion_control}

The main finding is the loss of an assumed relationship. When the detector is held fixed, better reconstruction fidelity does not imply better preservation of defect evidence. This is not a semantic distinction. If each reconstructed image is paired with a different detector smoothing parameter, thresholding rule, or calibration policy, the image transformation and scoring function change together. The resulting gain can no longer be attributed to super-resolution. In this paper the common-detector Track~A comparison removes that ambiguity.

The direct DeepLabV3 result has a different interpretation. It shows that the low-resolution observation contains more usable mask evidence than the common residual detector extracts under the present Track~A scoring rule. It does not show that DeepLabV3 is a reconstruction method, it is not capacity-matched to DPU-WaferSR, and it does not show that SR should be inserted before an existing detector. Keeping the direct predictor in Track~B makes the scientific claim sharper: reconstruction quality, direct detection, and joint reconstruction/detection are related but different experiments.

\subsection{Why reconstruction optimization can lose sparse defects}
\label{sec:discussion_reconstruction_loss}

The synthetic images contain large regular line/space and contact-hole backgrounds plus small defect regions. A reconstruction objective that is dominated by average pixel or structural similarity can improve the background while smoothing the residual signal that marks a bridge, gap, scratch, or particle. That tradeoff is visible in the trained reconstruction models: their SSIM improves, their FPR remains feasible, and their defect recall drops in every paired repetition relative to bicubic (\figref{fig:reconstruction_relationship}; \tabref{tab:paired_deltas}).

This behavior is consistent with a broader scientific-imaging concern: an image can look plausible while changing the measurement target. In inspection, plausibility is not enough because the downstream event is a thresholded call under a review budget. A method that preserves layout texture but attenuates a weak defect is not safer simply because it produces a visually cleaner image.

\subsection{Low false-call evaluation is policy transfer}
\label{sec:discussion_policy_transfer}

The DPU-WaferSR calibration result makes low-FPR evaluation a transfer problem. The candidate thresholds are fit on validation data, the final policy is selected on independent clean calibration images, and the held-out images are evaluated once. That policy passes clean calibration in all ten repetitions, but the held-out FPR exceeds the feasibility limit in all ten repetitions (\figref{fig:unified_protocol}; \tabref{tab:operating_transfer}; \figref{fig:calibration_transfer_appendix}). The conclusion is not that the threshold should be moved after seeing test labels; the conclusion is that clean-calibration feasibility did not transfer.

This is the practical reason to separate validation calibration, clean-policy selection, and held-out evaluation. A threshold is not only a number; it is an operating policy tied to a distribution of backgrounds, defect-adjacent structures, and score calibration. If the policy changes after test inspection, the study no longer measures transfer. Under this design, an unfavorable transfer result is evidence about the policy rather than a defect to tune away.

\subsection{External masks provide stress evidence, not production validation}
\label{sec:discussion_external}

Carinthia-S adds real SEM texture and public masks, which are valuable because they are outside the synthetic generator. It does not provide paired optical observations at two resolutions, and the pseudo-low-resolution formation used here is not a scanner model. The external layer is therefore interpreted as unchanged-policy stress evidence: do synthetic policies remain quiet, collapse, or over-activate when applied to real SEM texture?

Under that interpretation, the result is useful. The local reconstruction detector activates heavily on no-defect SEM crops, whereas DeepLabV3 transfers more quietly but still has seed-level instability (\figref{fig:external_transfer}; \tabref{tab:external_transfer}). This supports the same conclusion as the synthetic analysis: the operating policy and its transfer need to be measured directly.

\subsection{Implications for inspection-oriented super-resolution}
\label{sec:discussion_implications}

The benchmark suggests four requirements for future inspection-oriented SR studies. First, image quality should be reported with task-conditioned evidence metrics, not as a substitute for them. Second, clean-only controls should accompany defect-bearing images so recall gains can be separated from broad activation. Third, weak-defect and new-morphology sets should remain separate from model and threshold selection. Fourth, the original observation should remain available in the review path until task-conditioned validation shows that the reconstructed image preserves the relevant evidence.

The results also clarify how to use joint models such as DPU-WaferSR. A joint reconstruction/detection model can expose useful signals and provide a stress case for the protocol, but it should not be presented as a general architecture claim unless the comparison includes capacity-matched direct detectors, same-backbone no-SR ablations, task-trained SR baselines, and paired production validation. This manuscript uses DPU-WaferSR as an illustrative participant in the benchmark rather than as the main novelty.

\section{Limitations}
\label{sec:limitations}

The controlled synthetic layer is the source of exact masks, paired observations, weak-defect stress, held-out morphology, and clean controls. It is not a calibrated model of a specific scanner, process node, material stack, or fab acquisition recipe. The degradation model is explicit and reproducible, but paired production optical data would be required to validate scanner-specific conclusions.

The ten repeated units jointly vary image generation, model training, calibration samples, and threshold selection. They are appropriate for the end-to-end question asked here: whether the complete policy preserves evidence and transfers under the declared protocol. They do not decompose data-draw variance from training-seed variance. Larger future releases should include fixed-image repeated training seeds, additional independent generator draws, and hierarchical uncertainty estimates.

The method set is intentionally mixed. Track~A includes interpolation, classical restoration, deterministic enhancement, and compact learned reconstruction models under one common detector. It does not yet include a full set of task-trained EDSR, RCAN, SwinIR, HAT, diffusion, or SR-FABNet-style participants trained under exactly the same degradation and false-call protocol. Track~B includes a strong direct DeepLabV3 comparator and an illustrative joint model, but it is not a capacity-matched architecture tournament.

The primary feasibility rule is pixel-level FPR at a research target of $3\times10^{-4}$ with a $1.5\times$ held-out tolerance. Pixel FPR is important because it measures clean-region activation at the decision threshold, but it is not the whole review burden. Connected false components, field-level review routing, defect-class priority, tool throughput, and operator review time can change practical acceptance. Component, clean-control, edge, weak-defect, calibration, and external metrics are therefore reported as secondary evidence rather than hidden acceptance criteria.

Carinthia-S is public and useful for external stress, but it is highly imbalanced and unpaired with the synthetic optical-style acquisition. The external results should not be read as fab-scale performance claims. They show how unchanged synthetic policies behave on real SEM texture and identify transfer risk that paired production data should test more directly.

Finally, weak defects remain largely undetected by feasible policies. That is a scientific limitation of the present methods and a useful pressure point for the benchmark. A stronger future method should improve weak-defect evidence while keeping clean and nominal false calls within the declared operating budget.

\section{Conclusion}
\label{sec:conclusion}

WaferInspectSR-Bench evaluates super-resolution as an inspection-evidence problem rather than as a visual-quality problem. The controlled reconstruction track shows that learned SR can improve SSIM while losing defect recall under an identical downstream detector. The direct-detector track shows that substantially more nominal evidence can be recovered from the low-resolution observation when the prediction problem is trained for masks, subject to the capacity and architecture caveats in \secref{sec:limitations}. The joint DPU policy shows that independent clean calibration can fail to transfer to held-out images. The external Carinthia-S stress test shows that unchanged policies can shift sharply under real SEM texture.

The practical conclusion is narrow and actionable: inspection-oriented SR should be judged by preserved task evidence, clean-region silence, weak-defect sensitivity, and operating-policy transfer. A sharper image is not sufficient evidence that a defect decision is safer.

\section{Code and Data Availability}
\label{sec:availability}

The benchmark implementation, configuration files, and scripts needed to regenerate the reported experiments are available at \href{https://github.com/nbbllxx0/WAFERINSPECTSR-BENCH}{github.com/nbbllxx0/WAFERINSPECTSR-BENCH} under the MIT license. Carinthia-S images and masks must be obtained from their original source and remain subject to the dataset terms \cite{carinthias2025}.

\section*{Declaration of Generative AI and AI-Assisted Technologies}

During preparation of this manuscript, the authors used generative-AI tools for language editing, document formatting, and presentation-consistency checks. These tools were not treated as sources of scientific evidence and did not independently determine the study design, reported data, numerical results, citations, or conclusions. The authors checked all AI-assisted material against the underlying code, data, and cited sources, approved the final text and figures, and take full responsibility for the content.

%% file: sections/07_appendices.tex
\appendix

\section{Additional Protocol Details}
\label{app:protocol_details}

The main text gives the conceptual protocol. This appendix records enough detail for readers to reproduce the reported operating-point comparisons and to audit the split logic. The key design choice is that every decision rule is fixed before a held-out image is scored. Validation data fit temperatures and candidate thresholds, independent clean images choose among candidates, and held-out nominal, weak, morphology-shift, and external images are evaluated once.

\begin{algobox}{Synthetic sample generation and degradation}
\begin{enumerate}
  \item Draw a layout family, pitch, linewidth, local contrast, and background texture from the predeclared seed.
  \item Render a $256\times256$ high-resolution image and exact masks for defect, clean region, and edge structure.
  \item Insert one nominal defect from the bridge, gap, missing-pattern, particle, or scratch family; reserve residue-like defects for the held-out morphology split.
  \item Form the observation by applying blur, defocus, twofold downsampling, shot/Poisson/Gaussian noise, scan-line variation, contrast drift, and aliasing.
  \item Preserve masks and component labels across degradation; only the observation changes. The shared degradation configuration and per-sample seed identify and reproduce that draw.
\end{enumerate}
\end{algobox}

\begin{algobox}{Operating-point selection}
\begin{enumerate}
  \item Train or define the method using the training split only.
  \item Fit temperature scaling on validation-calibration samples when logits are available.
  \item For each predeclared candidate target, derive a threshold from validation-calibration clean-region scores.
  \item Apply each candidate threshold to independent clean-calibration samples.
  \item Select the most permissive candidate whose clean-calibration FPR is at most $4.5\times10^{-4}$.
  \item If no candidate is feasible, choose the most conservative candidate and record the policy as infeasible.
  \item Freeze model weights, temperature, threshold, prior weight, and candidate identity before scoring held-out images.
\end{enumerate}
\end{algobox}

\begin{algobox}{Held-out and external evaluation}
\begin{enumerate}
  \item Apply the fixed policy to nominal held-out, clean-only, weak-defect, and held-out morphology splits.
  \item Compute pixel recall, clean-region FPR, clean hallucination rate, precision, component recall, IoU-at-0.10 component recall, edge F1, reconstruction PSNR/SSIM, and calibration summaries.
  \item Aggregate image-level metrics within a repetition and then summarize the ten repetitions by mean and SD.
  \item For Carinthia-S, form pseudo-low-resolution SEM inputs, apply the unchanged synthetic policies, and report image-macro recall, class-macro recall, clean-mask FPR, and no-defect crop-call rate.
\end{enumerate}
\end{algobox}

\section{Model and Training Details}
\label{app:model_training}

\tabref{tab:appendix_training_details} summarizes implementation details that are needed to interpret the comparison. The reconstruction track intentionally uses compact models and simple transformations because its purpose is causal isolation of the image transformation under a fixed detector. The direct and joint predictors are reported separately because they optimize mask prediction directly. \tabref{tab:appendix_dpu_loss} identifies the five objectives used by the illustrative joint participant.

\begin{table}[!htbp]
\centering
\caption{Training and inference details for the main method families.}
\label{tab:appendix_training_details}
\small
\begin{tabularx}{\linewidth}{>{\raggedright\arraybackslash}p{0.20\linewidth}>{\raggedright\arraybackslash}p{0.23\linewidth}>{\raggedright\arraybackslash}X}
\toprule
Family & Learned parameters & Role in the comparison \\
\midrule
Interpolation and classical restoration & none & Non-learned transformations that establish how much defect evidence is preserved by simple image formation and local enhancement. \\
NAF heuristic & none & Deterministic residual enhancement that tests whether local contrast sharpening can help the common detector without training. \\
SR-lite and NAF trained & compact reconstruction networks & Learned reconstruction-only participants trained to reconstruct high-resolution images, then scored by the same local detector as every Track~A method. \\
DeepLabV3 & direct detector & Mask predictor trained on the low-resolution observation; used to test whether the observation contains task evidence when optimized for detection. \\
DPU-WaferSR & joint reconstruction, defect, and risk heads & Illustrative joint participant used to test calibration transfer and weak-defect behavior, not a capacity-matched architecture claim. \\
\bottomrule
\end{tabularx}
\end{table}

\begin{table}[!htbp]
\centering
\caption{Representative DPU-WaferSR loss terms. The model predicts a reconstruction, a defect probability map, and a risk map; the risk target is based on prediction error and is not used to select test thresholds.}
\label{tab:appendix_dpu_loss}
\small
\begin{tabular}{ll}
\toprule
Term & Purpose \\
\midrule
Reconstruction loss & Preserve high-resolution image content. \\
Positive-weighted defect loss & Mitigate sparse-positive imbalance in pixel mask prediction. \\
Clean-region penalty & Penalize activations in clean regions. \\
Edge consistency & Encourage boundary alignment near mask edges. \\
Risk loss & Predict residual error for risk-coverage analysis. \\
\bottomrule
\end{tabular}
\end{table}

\section{Full Synthetic Numeric Summaries}
\label{app:synthetic_tables}

The primary table in the main text gives the full method comparison. \tabref{tab:appendix_seed_synthetic} exposes seed-level values for representative reconstruction and direct-detector policies, while \tabref{tab:appendix_dpu_seed_selection} reports the DPU clean-calibration choice and held-out outcome for every seed. FPR columns marked as $10^{-4}$ use that scale to keep low false-call values readable.

\begin{table}[!htbp]
\centering
\caption{Seed-level held-out results for bicubic, NAF-trained SR, and DeepLabV3. FPR columns are in units of $10^{-4}$.}
\label{tab:appendix_seed_synthetic}
\scriptsize
\setlength{\tabcolsep}{4pt}
\begin{tabular}{rrrrrrr}
\toprule
Seed & Bicubic recall & Bicubic FPR & NAF recall & NAF FPR & DeepLab recall & DeepLab FPR \\
\midrule
7 & 0.0122 & 0.3080 & 0.0059 & 0.0801 & 0.1748 & 1.1889 \\
11 & 0.0206 & 0.6698 & 0.0077 & 0.6316 & 0.2134 & 1.7914 \\
13 & 0.0124 & 0.8089 & 0.0027 & 0.0902 & 0.1533 & 1.7834 \\
17 & 0.0112 & 0.4228 & 0.0026 & 0.0128 & 0.2183 & 3.2486 \\
19 & 0.0042 & 0.3111 & 0.0024 & 0.0449 & 0.2410 & 1.4629 \\
23 & 0.0181 & 0.8446 & 0.0051 & 2.3266 & 0.2742 & 1.4976 \\
29 & 0.0172 & 1.0088 & 0.0009 & 0.4489 & 0.1533 & 0.8170 \\
31 & 0.0069 & 0.5458 & 0.0018 & 0.0802 & 0.1535 & 0.8934 \\
37 & 0.0121 & 0.7687 & 0.0089 & 0.2784 & 0.2016 & 1.3695 \\
41 & 0.0108 & 0.3484 & 0.0058 & 0.2711 & 0.2007 & 3.3788 \\
\bottomrule
\end{tabular}
\end{table}

\begin{table}[!htbp]
\centering
\caption{DPU-WaferSR clean-calibration selection and held-out outcome by seed. Clean-calibration and test FPR columns are in units of $10^{-4}$; weak recall is in units of $10^{-3}$.}
\label{tab:appendix_dpu_seed_selection}
\scriptsize
\setlength{\tabcolsep}{4pt}
\begin{tabular}{rrrrrrr}
\toprule
Seed & Candidate target & Threshold & Clean-calib FPR & Recall & Test FPR & Weak recall \\
\midrule
7 & $0.00015$ & $0.863$ & $3.63$ & $0.0671$ & $13.40$ & $0.488$ \\
11 & $0.0003$ & $0.868$ & $2.34$ & $0.0921$ & $21.34$ & $10.116$ \\
13 & $0.00015$ & $0.891$ & $4.36$ & $0.0487$ & $7.97$ & $0.790$ \\
17 & $0.0001$ & $0.820$ & $4.04$ & $0.0805$ & $19.19$ & $6.977$ \\
19 & $0.0003$ & $0.814$ & $1.80$ & $0.0504$ & $29.44$ & $2.598$ \\
23 & $0.0003$ & $0.822$ & $3.92$ & $0.0543$ & $12.60$ & $2.622$ \\
29 & $0.0003$ & $0.756$ & $3.87$ & $0.1016$ & $11.18$ & $1.457$ \\
31 & $0.00015$ & $0.866$ & $4.43$ & $0.0550$ & $22.53$ & $0.340$ \\
37 & $0.0002$ & $0.810$ & $3.40$ & $0.0811$ & $12.95$ & $3.822$ \\
41 & $0.0003$ & $0.862$ & $3.95$ & $0.0596$ & $14.34$ & $3.319$ \\
\bottomrule
\end{tabular}
\end{table}

\begin{figure}[!htbp]
  \centering
  \includegraphics[width=\linewidth]{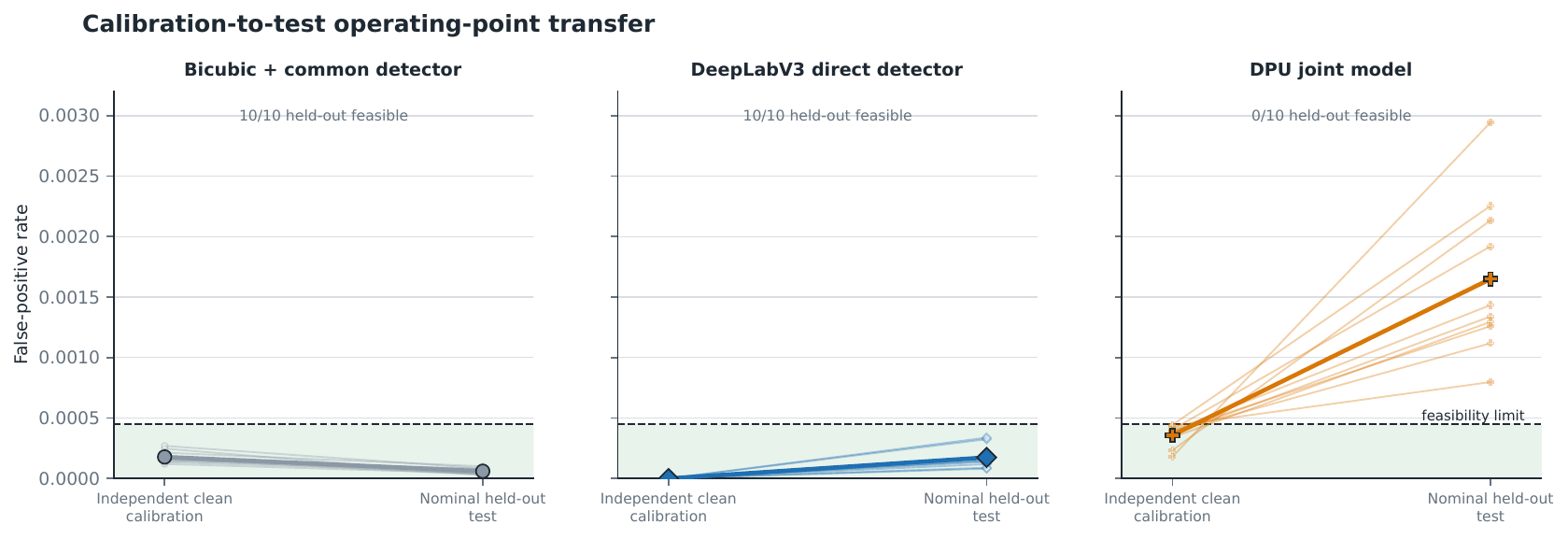}
  \caption{Seed-paired operating-point transfer from independent clean calibration to nominal held-out testing. Thin lines pair the same seed and thick lines connect method means. The shaded band and dashed boundary mark FPR values at or below $4.5\times10^{-4}$. All panels use the same linear vertical scale.}
  \label{fig:calibration_transfer_appendix}
\end{figure}

\figref{fig:calibration_transfer_appendix} shows the seed-level calibration-to-test movement directly. Bicubic and DeepLabV3 remain below the held-out feasibility limit in all ten repetitions. Every DPU policy is feasible on independent clean calibration images, yet every paired held-out endpoint lies above the same limit. The shared linear scale preserves the magnitude of this transfer gap. \tabref{tab:appendix_clean_counts} provides pooled clean-pixel counts and Wilson intervals as descriptive checks alongside the seed-level analysis.

\begin{table}[!htbp]
\centering
\caption{Selected clean-pixel count checks. Aggregate FPR is computed from summed false pixels and clean pixels, so it can differ slightly from the mean of seed-level FPR values.}
\label{tab:appendix_clean_counts}
\small
\begin{tabular}{llrrrr}
\toprule
Method & Split & Clean pixels & False pixels & Aggregate FPR & Wilson 95\% CI \\
\midrule
Bicubic & clean calibration & 10,485,760 & 1,870 & 0.000178 & [0.000170, 0.000187] \\
Bicubic & nominal test & 31,171,952 & 1,882 & 0.000060 & [0.000058, 0.000063] \\
NAF trained & nominal test & 31,171,952 & 1,329 & 0.000043 & [0.000040, 0.000045] \\
SR-lite & nominal test & 31,171,952 & 606 & 0.000019 & [0.000018, 0.000021] \\
DeepLabV3 & nominal test & 31,171,952 & 5,433 & 0.000174 & [0.000170, 0.000179] \\
DPU-WaferSR & nominal test & 31,171,952 & 51,426 & 0.001650 & [0.001636, 0.001664] \\
\bottomrule
\end{tabular}
\end{table}

\section{External Transfer Details}
\label{app:external_tables}

\tabref{tab:appendix_external_seed} reports seed-level external values for the three policies shown in the main external table. The external layer uses image-macro and class-macro summaries because the six Carinthia-S classes are highly imbalanced; class 3 dominates the sample count, while classes 2 and 5 contain only 8 and 4 masks, respectively. \tabref{tab:appendix_carinthia_counts} reports the complete class distribution.

\begin{table}[!htbp]
\centering
\caption{Seed-level Carinthia-S transfer. FPR columns are in units of $10^{-4}$.}
\label{tab:appendix_external_seed}
\scriptsize
\setlength{\tabcolsep}{4pt}
\begin{tabular}{rrrrrrr}
\toprule
Seed & Bicubic recall & Bicubic FPR & NAF recall & NAF FPR & DeepLab recall & DeepLab FPR \\
\midrule
7 & 0.0132 & 86.52 & 0.0234 & 0.85 & 0.2218 & 3.41 \\
11 & 0.0183 & 98.55 & 0.0149 & 5.41 & 0.3116 & 4.79 \\
13 & 0.0185 & 98.78 & 0.0151 & 64.07 & 0.0069 & 0.01 \\
17 & 0.0139 & 89.61 & 0.0385 & 69.94 & 0.3286 & 4.71 \\
19 & 0.0160 & 95.91 & 0.0187 & 17.55 & 0.1067 & 1.19 \\
23 & 0.0177 & 98.04 & 0.0060 & 39.29 & 0.1481 & 0.92 \\
29 & 0.0170 & 97.30 & 0.0047 & 95.65 & 0.1425 & 4.64 \\
31 & 0.0137 & 88.43 & 0.0179 & 2.75 & 0.0894 & 0.14 \\
37 & 0.0184 & 98.72 & 0.0251 & 6.30 & 0.0829 & 0.48 \\
41 & 0.0118 & 82.98 & 0.0107 & 35.19 & 0.0073 & 0.02 \\
\bottomrule
\end{tabular}
\end{table}

\begin{table}[!htbp]
\centering
\caption{Carinthia-S class counts used for external stress reporting. Source class names follow the public dataset metadata; the counts are used only for reporting and never for threshold selection.}
\label{tab:appendix_carinthia_counts}
\small
\begin{tabular}{lrr}
\toprule
Class id & Samples & Share \\
\midrule
1 & 55 & 1.20\% \\
2 & 8 & 0.17\% \\
3 & 4,008 & 87.30\% \\
4 & 289 & 6.30\% \\
5 & 4 & 0.09\% \\
6 & 227 & 4.94\% \\
\bottomrule
\end{tabular}
\end{table}

\section{Metric Definitions}
\label{app:metric_definitions}

Pixel recall is the fraction of defect-mask pixels called positive. Clean-region FPR is the fraction of clean-mask pixels called positive. Clean hallucination rate is measured on clean-only controls and uses the same clean-region denominator. Precision is the fraction of called pixels that overlap the defect mask. Component recall counts a ground-truth component as detected when any predicted component overlaps it, while IoU-at-0.10 component recall requires at least 0.10 intersection-over-union. Edge F1 measures overlap between predicted and reference edge bands. Image-macro external recall averages recall by image before aggregating, and class-macro external recall averages classes 1--5 after within-class averaging (class 6 is no-defect and is reported through clean-mask FPR / crop-call rate instead).

The feasibility rule is applied to nominal held-out clean-region FPR. The target is $3\times10^{-4}$ and the accepted research tolerance is $4.5\times10^{-4}$. The tolerance is used only for deciding whether a policy remains feasible; all raw FPR values are reported.